\documentclass[12pt]{article}
\usepackage[top=1in,bottom=1in,left=1in,right=1in]{geometry}
\usepackage[T1]{fontenc}

\usepackage[
  backend=biber,
  style=nature,
  sorting=none,
  sortcites=true,
  autocite=superscript,
  natbib=true
]{biblatex}
\newsavebox{\tabbox}
\newlength{\tabnotew}
\newcommand{\settabnotew}{\setlength{\tabnotew}{\wd\tabbox}\ifdim\tabnotew>\textwidth\setlength{\tabnotew}{\textwidth}\fi}
\usepackage[bottom]{footmisc}
\usepackage{caption}
\usepackage{ragged2e}
\usepackage{mathpazo}
\usepackage{amsmath, amsfonts}
\usepackage{graphicx}
\usepackage{graphbox}
\usepackage[update,prepend]{epstopdf}
\usepackage{float}
\usepackage{booktabs}
\usepackage{multirow}
\usepackage{array}
\usepackage{tabularx}
\newcolumntype{L}[1]{>{\raggedright\arraybackslash}p{#1}}
\usepackage{xcolor}
\definecolor{myLightGray}{RGB}{191,191,191}
\definecolor{myGray}{RGB}{160,160,160}
\definecolor{myDarkGray}{RGB}{144,144,144}
\definecolor{myDarkRed}{RGB}{167,114,115}
\definecolor{myRed}{RGB}{255,58,70}
\definecolor{myGreen}{RGB}{0,255,71}
\definecolor{headerblue}{RGB}{43, 87, 151}
\definecolor{lightgray}{RGB}{220, 230, 241}
\definecolor{inputblue}{RGB}{70,130,180}
\definecolor{featgreen}{RGB}{76,153,100}
\definecolor{axisyellow}{RGB}{210,170,60}
\definecolor{threshred}{RGB}{190,80,80}
\definecolor{trainpurple}{RGB}{120,80,160}
\definecolor{outputgray}{RGB}{90,90,90}
\definecolor{lightbg}{RGB}{245,245,250}
\definecolor{annotgray}{RGB}{100,100,100}
\definecolor{darkgr}{RGB}{60,60,60}
\definecolor{medgr}{RGB}{100,100,100}
\definecolor{litegr}{RGB}{180,180,180}
\definecolor{skipgr}{RGB}{120,120,120}
\usepackage{setspace}
\usepackage{pdflscape}
\usepackage{fancyhdr}
\usepackage[colorlinks, citecolor=black, linkcolor=black, urlcolor=black]{hyperref}
\usepackage{chngcntr}

\usepackage{tikz}
\usetikzlibrary{positioning, arrows.meta, calc, fit, backgrounds, shapes.geometric, decorations.pathreplacing, calligraphy, arrows}
\usepackage{pgfplots}
\pgfplotsset{compat=1.18}
\usepgfplotslibrary{groupplots, fillbetween}
\definecolor{panelblue}{RGB}{31,78,121}
\definecolor{panelorange}{RGB}{209,116,31}
\definecolor{panelgreen}{RGB}{46,139,87}
\definecolor{panelred}{RGB}{192,80,77}
\definecolor{panelpurple}{RGB}{122,59,143}
\definecolor{panellight}{RGB}{143,169,196}
\tikzstyle{bag} = [align=center]
\tikzset{
  state/.style={
    rectangle,
    rounded corners,
    draw=black, very thick,
    minimum height=2em,
    inner sep=2pt,
    text centered,
  },
}

\title{Human versus Computer Vision}
\begin{document}
\author{Elena Sirotkina\\Center for Data Science, New York University}

\date{}

\maketitle

\begin{abstract}
\singlespacing \noindent
Computer vision saliency models predict where people will look, one map per image, and a predicted-attention industry sells those maps in place of measuring real viewers \autocite{threeM_vas, jiang2023ueyes, twitter2021cropping}. I evaluate the leading models against every publicly available eye-tracking corpus that records viewer demographics, comprising 27{,}384 participants across eight datasets, including 11.4 million webcam gaze points from 2{,}993 US adults recruited to national quotas while viewing circulating news images \autocite{linka2025protracted, strauch2023saliency, alexander2017open}. I show that a single map predicts where every group looks about equally well, but systematically ignores how long each place holds a group’s attention. That time follows a clear sociodemographic and political fingerprint, so the map carries a systematic bias against the youngest and oldest viewers and against Black, lower-income, and Republican-leaning ones. I use each group’s own gaze to test whether it carries a learnable pattern, across every demographic axis the sample supports. Showing which audiences these systems can learn and what determines this sets the standard for judging claims that computer vision captures diverse human vision.
\end{abstract}

\newpage
\doublespacing

We increasingly live in a world where computational systems imitate human behavior. A computer vision model of visual attention, called a saliency model, takes an image and returns a single map that assigns every location a number for how likely a viewer is to look there. Since the approach was first formalized \autocite{itti1998model}, that map has moved out of the laboratory and into commercial and institutional decisions about what people see. Advertisers read a predicted-attention map in place of a panel of eye-tracked shoppers \autocite{threeM_vas}, interface teams choose layouts by where a model says the eye will land \autocite{jiang2023ueyes}, driver-assistance and clinical pipelines weight a scene toward where a driver or a radiologist would look \autocite{palazzi2019dreyeve, karargyris2021creation}, and social platforms have cropped billions of image previews to a model's single most salient point \autocite{twitter2021cropping}. In every one of these uses the model stands in for a human observer, with one map per image, identical for everyone who looks at it.

That single map carries two claims. The first is that the content of a real, circulating image predicts where people look. The second is that one prediction serves every viewer, so a model trained on some people's attention counts as a model of human attention. Both rest on laboratory benchmarks. Those benchmarks know their observers as a group or not at all. The main laboratory corpus recorded 15 young adults \autocite{judd2009learning}, and the largest follows the mouse of crowd workers over a blurred image in place of their eyes \autocite{jiang2015salicon}. A benchmark can therefore test a model only against the pool. To learn whom a model serves and whom it fails, it has to be tested on viewers whose characteristics are known, looking at the kinds of images these maps are used on \autocite{strauch2023saliency}. 

Social science gives at least two reasons to question the second claim, that one prediction serves every viewer. Where people look differs from one person to another, and these differences persist over time and reflect what an image means to the viewer \autocite{dehaas2019individual, xu2019personalized}. They also should vary systematically across groups. Culture \autocite{chua2005cultural}, gender, and age \autocite{linka2025protracted, sulflow2019selective} shape visual attention, and in political settings eye tracking shows selective attention to
news and links gaze to subsequent voting decisions
\autocite{ jenke2021using}.

I test both claims on all publicly available eye-tracking corpora that record viewers’ characteristics, 27{,}384 viewers aged 3 to 95 (Methods). They include the national sample of this study, 2{,}993 US adults recruited to national quotas who viewed 83 news images at home while a webcam recorded their gaze \autocite{papoutsaki2016webgazer}, and seven corpora from museums and laboratories \autocite{linka2025protracted, cenek2025cross, wilming2017extensive, strauch2023saliency, sumer2021fakenewsperception, alexander2017open, griffith2021gazebase}. In every corpus I measure how groups differ in where they look and in how long they look, and how much of each difference a single map misses.

Deep networks predict where each group looks about equally well, because the
groups tend to look at much the same places. The models leave out the time per
fixation, which measures how much attention a place receives and which previous research has linked to subsequent memory for the viewed location \autocite{wedel2000eye}. This time
differs between groups by up to 36\%, and I report the
difference as the bias of the model (Table~\ref{tab:rest_time}, Extended Data Table~\ref{tab:rest_time_all}).  Age gives the largest bias. The oldest viewers spend 6 to 25\% more time per fixation than young adults across the museum and the national sample, and young children fixate longer as well and give text under a third of the time young adults give it (Fig.~\ref{fig:outside_map}). Race gives a bias of 12\% in two very different samples, with Black viewers spending more time per fixation than White viewers both on a laboratory tracker and on a webcam at home. Lower-income viewers spend 8\% more time per fixation than upper-income viewers, and on news images on four political topics viewers whose positions lean Republican spend 20\% more than those whose positions lean Democratic.

I measure what this bias costs in a simulation that cuts every recorded viewing
at 1.7~seconds, the time a post in a social media feed holds the eye on a phone
\autocite{facebook2016capturing}. In that time the groups the map is biased
against reach 8 to 17\% fewer of the five places the single map ranks first
(Fig.~\ref{fig:feed}). In a guided-viewing experiment, moving fixations away from the
informative places of a picture lowered its later recall by about a fifth \autocite{fehlmann2020visual}, and
errors in political information that a whole group shares persist in the
election outcome \autocite{bartels1996uninformed}. If memory follows fixations here as it does in previous studies, groups disadvantaged by the map may miss information that others receive from the same content, which could systematically shape what different groups know and believe.

The same data point to a remedy. Where people look depends largely on the image, while how long they look depends more on the viewer \autocite{nuthmann2010crisp, tatler2017latest}. Since groups differ mainly in fixation duration, correcting the bias requires greater attention to how long viewers look. The comparison
between human and computer vision depends on which humans enter it. I test whether such a correction is possible in two steps. First, a classifier tests whether fixation durations contain enough information to distinguish viewers by social group. Second, a neural network tests whether information about the viewer’s group helps predict where and how long they look. These results suggest that audience-specific attention models could reduce the systematic bias in attention maps used by downstream applications. And by accounting for whose attention these maps represent, such models could help prevent unequal access to visual information that arises when systems consistently overlook how some groups view the same content.


\section*{Results}
\label{sec:results}

\subsection*{Design and measurement}

The analyses use nine corpora (Methods, Extended Data Table~\ref{tab:sample_composition}, Supplementary Table~\ref{tab:datasets}). The national sample of this study holds 2{,}993 US adults recruited to national quotas, who viewed 83 news images at home while a webcam recorded their gaze. Two museum corpora record visitors on installed trackers, that of Linka and colleagues with 12{,}056 visitors aged 3 to 95 viewing everyday scenes \autocite{linka2025protracted} and that of Strauch and colleagues with about 8{,}000 visitors viewing one image \autocite{strauch2023saliency}. The students of \v{C}en\v{e}k and colleagues come from seven samples in six countries \autocite{cenek2025cross}, and the laboratory data of Wilming and colleagues from 58 viewers in three age groups \autocite{wilming2017extensive}. Children and adolescents of the Healthy Brain Network \autocite{alexander2017open} and the university students of GazeBase \autocite{griffith2021gazebase} watched short videos, the FakeNewsPerception corpus records viewers of news pages \autocite{sumer2021fakenewsperception}, and MIT1003 is the laboratory benchmark of 15 young adults that trained DeepGazeIIE and TranSalNet \autocite{judd2009learning}. Each corpus records different characteristics of its viewers, and together they span ages 3 to 95 and the social and political characteristics of Extended Data Table~\ref{tab:sample_composition}.  Every group comparison stays within one corpus, on the same images and the same instrument, and the same comparison repeated across corpora shows whether a difference holds across settings and instruments. Extended Data Table~\ref{tab:spec_grid} and Methods repeat the national comparisons with each viewer's recording error and recording quality in the model, and Extended Data Table~\ref{tab:rest_time_all} repeats the museum age comparison on visitors calibrated within one degree.

\begin{figure}[!ht]
\centering
\caption{The groups share the map’s places and part in how they look.}
\label{fig:method_pipeline}
\includegraphics[width=\textwidth]{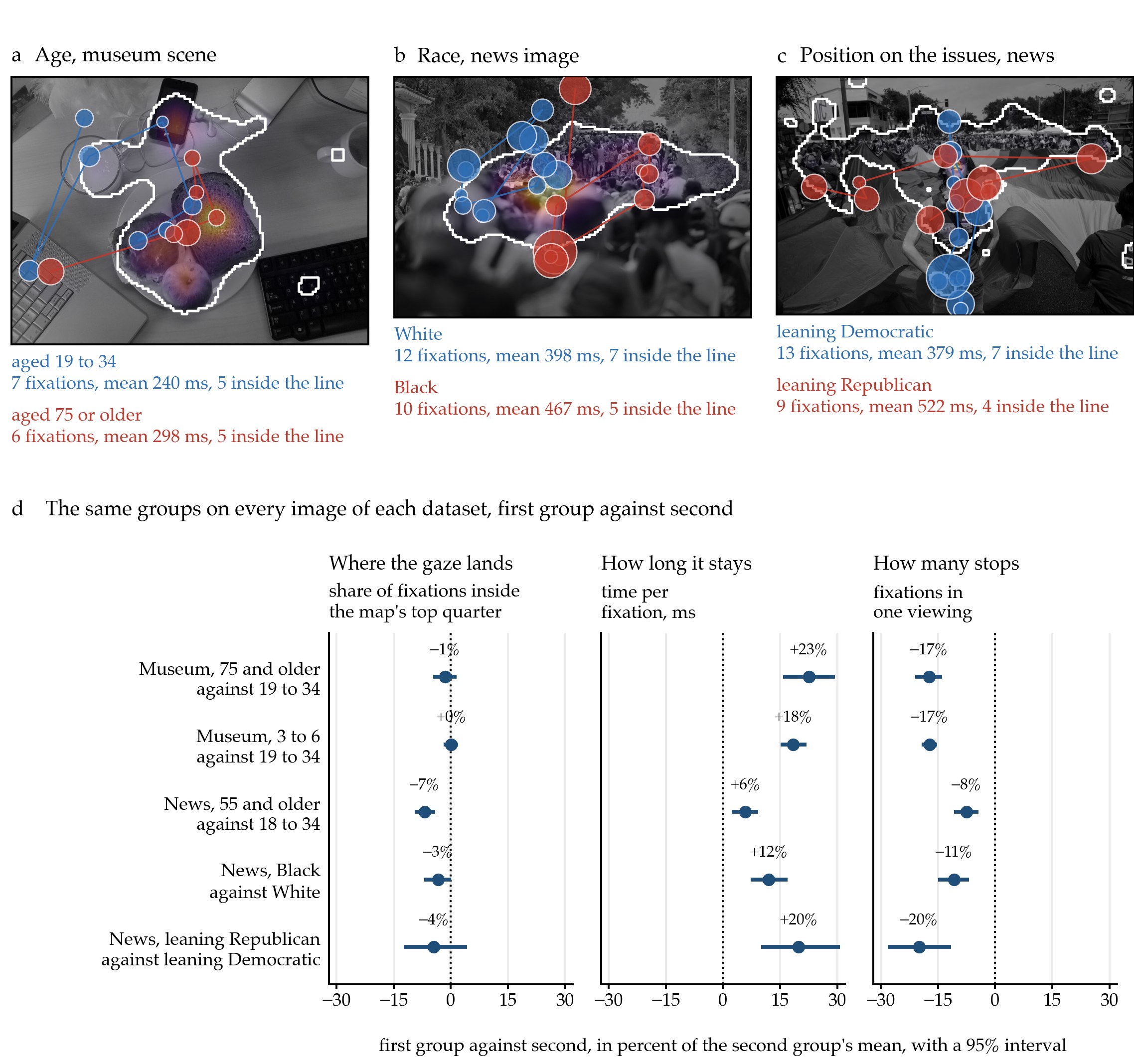}
\par\smallskip
{\scriptsize\justifying\textit{Notes.} \textbf{a} to \textbf{c}, the map of DeepGazeIIE, a deep network that predicts gaze from the image after training on recorded human attention \autocite{linardos2021deepgaze}, is shaded, and the white line encloses the quarter of the image it ranks highest, the same for every viewer. Each image carries one typical viewer from each group, the viewer closest to the group’s medians on that image, the second group of panel d in blue and the first in red. A circle is one fixation, with its area proportional to its duration, and the line joins the fixations in order. \textbf{d}, differences between group means in percent of the second group’s mean, with 95\% intervals from 2,000 bootstrap draws of viewers, after removing the constant of each image and, on the news images, of the device. Extended Data Tables~\ref{tab:rest_time_all} and~\ref{tab:places_text} give the group sizes, Methods defines a fixation and the time per fixation, and Extended Data Table~\ref{tab:places_text} defines the number of fixations.\par}
\end{figure}

Throughout the study, I compare what a saliency map predicts with what groups of viewers do on the same image in two respects, where they look and how long each look lasts. Every tracker returns a stream of gaze positions, from which I extract fixations, the pauses in which the eyes stay in one place. In Figure~\ref{fig:method_pipeline} the shading is the map of the deep network DeepGazeIIE \autocite{linardos2021deepgaze}, and the white line encloses the quarter the map ranks highest. Each image shows two viewers in blue and red, each circle a fixation sized by its duration, joined in order. On the museum scene (panel a) a visitor aged 19 to 34 and a visitor aged 75 or older each put five fixations inside the line, but the older visitor makes six fixations averaging 298~ms against seven averaging 240~ms. The news images repeat this for a White and a Black viewer (panel b), 12 fixations of 398~ms against 10 of 467~ms, and for viewers leaning Democratic and Republican (panel c), 13 of 379~ms against 9 of 522~ms. In every pair the red viewer stops fewer times and stays longer at each stop.

Panel d shows that these viewers are typical of their groups. Across every image of each dataset, the groups put nearly the same share of their fixations inside the map's top quarter, within 7\% of the second group's share, while they differ by 6 to 23\% in the time per fixation and by 8 to 20\% in the number of fixations, each in percent of the second group's value with 95\% intervals in the panel. On the webcam the time per fixation runs from the onset of one fixation to the onset of the next, so there the two measures divide the same five seconds and show one difference. A single map therefore covers the places each group looks at about
equally well and misses how long each group looks. The size of that difference is the bias of the model.

To test how well a map predicts where people look, I score it against recorded gaze (Extended Data Fig.~\ref{fig:score_pipeline}). The score is the area under the curve (AUC), the probability that the map ranks a place a viewer fixated above a comparison place, so 0.5 is chance (panel c). Comparison places drawn uniformly across the frame give the raw score, which asks whether the map finds where people look. Comparison places drawn from gaze on other images carry the pull toward the center that all viewers share, so this shuffled area under the curve \autocite{bylinskii2019different}, which I call the center-corrected score, asks what the map knows about the image itself. As a human reference, a map built from one panel of viewers of an image is scored on viewers held out of it (Methods).

\subsection*{What the model families compute}

The six models tested here represent what is available to anyone predicting attention today. Four are deep networks trained on recorded human attention and distributed with released weights \autocite{linardos2021deepgaze, lou2022transalnet, droste2020unisal}. Two are classical detectors that compute a map from image statistics alone, with human gaze left out of their construction \autocite{hou2007saliency, montabone2010human}.

The two families turn out to measure different things, and regressing each map on four classical predictors of attention shows what each one measures (Extended Data Table~\ref{tab:saliency_predictors}). The deep networks are driven by distance from the center of the frame, and the classical detectors by local differences in brightness. The deep networks learned from recordings of human attention, and people attend near the middle of almost any image, so the networks took on that habit. The classical detectors never saw human attention and compute contrast by design. The four networks, which were trained on recordings of human attention, come closer to real viewers than the detectors, as the next subsection shows.

All six models share one property. Each returns a single map for an image, the same for whoever looks at it, so differences in attention between viewers stay invisible to it.

\subsection*{How much a model adds beyond the center}

The first question is how much the content of an image tells us about where people will look. Viewers look near the middle of the frame whatever the image shows \autocite{tatler2007central, tseng2009quantifying}, and at home this habit does most of the work, so a blank map that marks the center ranks more pairs correctly on the raw score than any of the six models (Table~\ref{tab:raw_auc}). The webcam reads the eye coarsely, yet the center map keeps its lead even among the tenth of viewers the webcam recorded most precisely (Extended Data Fig.~\ref{fig:measurement_error}a). The lead therefore reflects how people look at news on their own screens.

\begin{table}[!ht]
\centering
\caption{Removing the shared central habit turns the ranking over.}
\label{tab:raw_auc}
\small
\sbox{\tabbox}{\begin{tabular}{@{}l rr r@{}}
\toprule
& \multicolumn{2}{c}{News images (webcam)} & MIT1003 \\
\cmidrule(lr){2-3}
Predictor & Raw & Center-corrected & (laboratory) \\
\midrule
Central Gaussian & 0.715 & 0.502 & 0.828 \\
DeepGazeIIE & 0.692 & 0.534 & 0.905 \\
TranSalNet-Dense & 0.689 & 0.541 & 0.881 \\
TranSalNet-Res & 0.688 & 0.540 & 0.879 \\
UNISAL & 0.681 & 0.543 & 0.878 \\
Spectral Residual & 0.563 & 0.522 & 0.678 \\
Fine-Grained & 0.549 & 0.513 & 0.643 \\
Held-out audience of the image & \textbf{0.735} & \textbf{0.547} & \textbf{0.912} \\
\midrule
Share of gaze in the central circle & 0.461 & & 0.674 \\\bottomrule
\end{tabular}}%
\usebox{\tabbox}
\par\smallskip
{\scriptsize\justifying\settabnotew\noindent\parbox{\tabnotew}{\setlength{\emergencystretch}{2em}\textit{Notes.} Agreement in correct rankings, so 0.5 is chance. The central Gaussian is the center map of the text. MIT1003 is the laboratory benchmark of 15 young adults viewing natural scenes \autocite{judd2009learning}. The raw score draws its comparison point uniformly in the frame and the center-corrected score draws it from gaze on the other 82~images, and both score the maps as released. The held-out audience is a map built from the gaze of other viewers of the same image and scored on viewers whose gaze stayed out of it. On the center-corrected column the deep networks stand 0.033~to~0.041 above the center map and the classical detectors 0.011~to~0.020 above it. The central circle has a radius of 0.25 of the frame, and its row is a share of recorded gaze, so the center-corrected column is left empty there.\par}}
\end{table}

Once the center-corrected score removes this shared pull toward the middle, every model ranks above the center map (Extended Data Table~\ref{tab:center_increment}a). To learn how much an image can add, I replace the model with real viewers and predict each viewer from a map built from 256 others who saw the same image (Methods). That map gains about five correct rankings in 100 over the center map, and the deep networks recover seven tenths to all of this gain, while the classical detectors recover far less (Extended Data Table~\ref{tab:center_increment}b). So the networks already predict where people look nearly as well as a map built from 256 viewers. Any group differences that emerge therefore cannot simply be attributed to poor overall prediction accuracy and may reflect the limitations of using the same attention map for everyone.

The coarse reading of the webcam would matter for these comparisons only if it fell unevenly on the groups. With each viewer's own error in the model, the gaps by age and by political extremity stay well clear of chance, and the race gap holds at $p<0.05$ (specifications S5 and S6 of Extended Data Table~\ref{tab:spec_grid}). The comparisons below therefore set groups against each other within the same recordings and score the maps on the center-corrected score.

\subsection*{Whom the single map underrepresents}

\begin{figure}[!htbp]
\centering
\caption{A map built for one group adds little to the map for everyone.}
\label{fig:map_agree}
\includegraphics[width=0.87\textwidth]{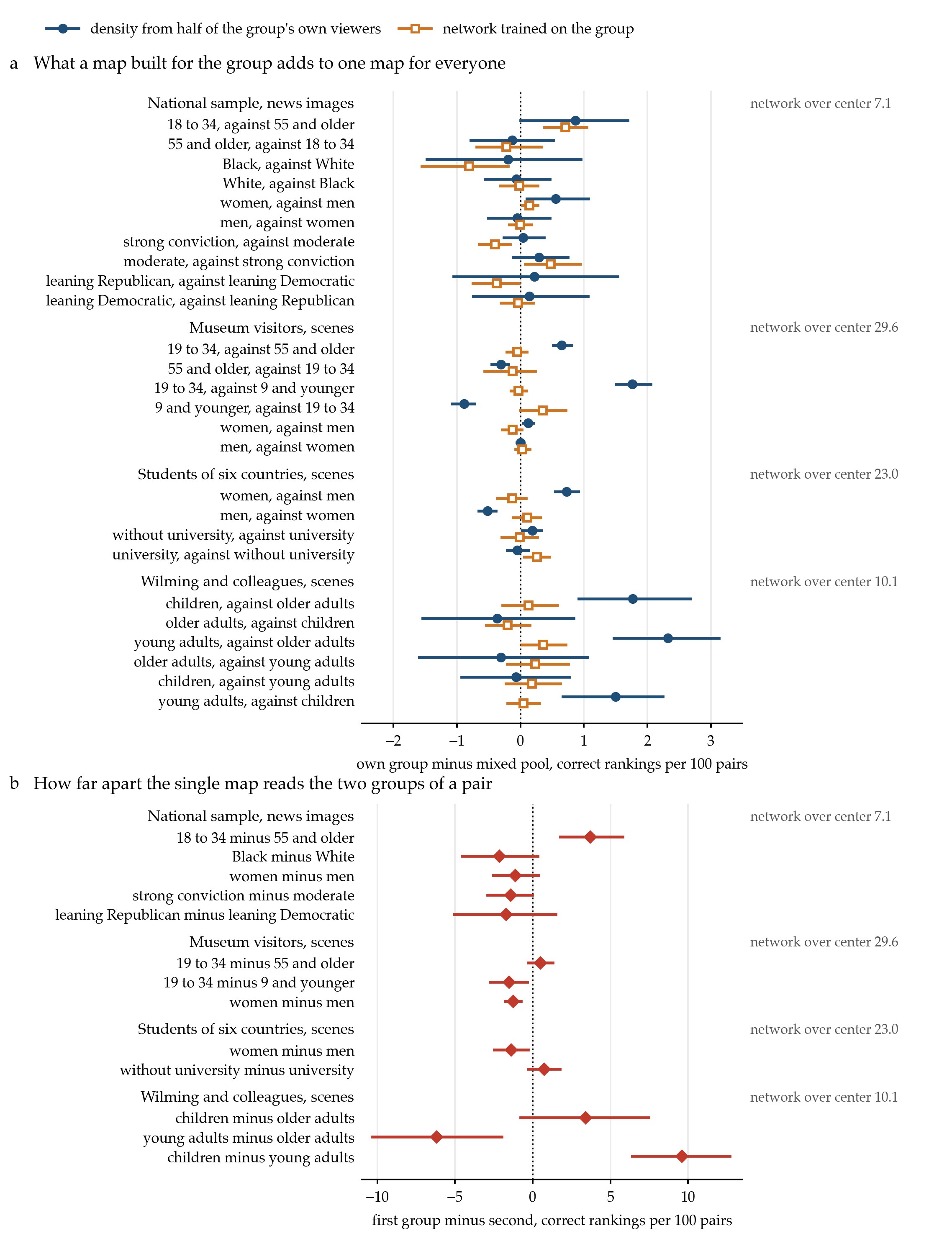}
\par\smallskip
{\footnotesize\justifying\textit{Notes.} Correct rankings per 100 pairs on the center-corrected score, on images held out of every fit, with 95\% intervals from 1,000 bootstrap draws. \textbf{a}, the gain for a group from a map built from its own viewers over a map built from a mixed pool of the same size, for a density of the viewers’ gaze and for a network trained on that gaze. \textbf{b}, the accuracy of the released UNISAL map for the first group of each pair minus its accuracy for the second, read against the gain of UNISAL over the center map, which stands at the right of each dataset. The gaps of the national sample in panel b are raw means over images, and Extended Data Table~\ref{tab:spec_grid} gives the adjusted estimates.\par}
\end{figure}

A map can serve a group badly in two ways. It can point to places the group does not look at, or it can be wrong about how long the group looks at each place. Older and younger viewers, Black and White viewers, and viewers at different positions on the political scale look at largely the same parts of an image, and the map locates these parts with about the same accuracy for each group (Fig.~\ref{fig:method_pipeline}d, Extended Data Table~\ref{tab:public_models}). The differences that remain in the national sample are small (Extended Data Table~\ref{tab:spec_grid}). A map built only from the gaze of one group predicts that group only marginally better than a map built from all viewers (Fig.~\ref{fig:map_agree}, Extended Data Table~\ref{tab:group_readouts}). The single map therefore treats groups unequally mostly through how long they look.

The groups differ substantively in how long they look at each place, so the model is wrong here. A saliency map ranks the places of an image the same way for every viewer, and a layout that follows the map assumes that every viewer reaches its top places. A viewer who stays longer at each place reaches fewer places in the few seconds people spend on an image, so the map overstates what slower groups see. The difference in time per fixation shows the size of this bias (Table~\ref{tab:rest_time}), and the feed simulation below measures it directly.

In the museum, visitors over 75 fixate 23\% longer than visitors aged 19 to 34 and make 17\% fewer fixations, so they cover a smaller part of each scene \autocite{linka2025protracted}. Young children fixate 16 to 36\% longer than adolescents or young adults in three corpora. Among webcam viewers, those aged 55 and older spend 6\% more time per fixation than those aged 18 to 34 on the full set of news images and 25\% more on the images of four political topics. Age changes the time per fixation more than any other characteristic and in more corpora. The difference persists at 18\% among museum visitors over 75 who wore no glasses and were calibrated within one degree, and at 4 to 6\% on the webcam once recording quality is held constant (Methods), which places its source in the viewers themselves. The single exception is a laboratory study in which older adults judged after each scene whether a small patch had appeared in it and fixated slightly shorter than young adults \autocite{acik2010developmental, wilming2017extensive}.

Race was recorded in two samples that differ in instrument, material and the age of the viewers. In both, Black viewers spend 12\% more time per fixation than White viewers and make 11\% fewer fixations. The estimate among university students watching videos on a laboratory tracker rests on 29 students who gave Black or African American as their only ethnicity \autocite{griffith2021gazebase}, while the estimate among adults viewing news images through a webcam rests on 335 and remains at 7 to 8\% when recording quality is held constant (Methods). The two samples measure the time per fixation in different ways (Methods) and give the same 12\%, which makes this one of the most consistent results of the study.

Political positions show some of the largest differences in fixation duration, although the evidence comes from the national sample alone. On news images covering four political topics, viewers whose positions lean Republican spend 20\% more time per fixation and make 20\% fewer fixations than those leaning Democratic, with the duration gap remaining after demographic controls (Methods). The gap falls to 2\% on immigration images alone, close to the 2\% difference by party identification, and the age gap is also larger on the images covering four topics. These differences therefore vary substantially across the sets of news images. Income shows a smaller difference of 8\% in the national sample but less than 1\% among students grouped by family class, whereas sex and education show no consistent direction across corpora (Extended Data Table~\ref{tab:rest_time_all}).

In the museum, children aged 3 to 6 spend 6.4\% of their fixation time on the text of a scene, against 21.6\% for young adults. The networks assign text 13.6\% of the map for all viewers, twice the share the children give it and a third less than the share young adults give it. On the news images, lower-income viewers reach the text in 29\% fewer viewings than upper-income viewers, and Black viewers in 17\% fewer than White viewers (Fig.~\ref{fig:outside_map}, Extended Data Table~\ref{tab:places_text}). Among adults, text thus shows most clearly how the slower pace of a group translates into less of the image seen; most children aged 3 to 6 cannot yet read.

A regression on the characteristics that viewers report predicts their time per fixation. In the national sample, the fifth of held-out viewers predicted to be slowest spend 17\% more time per fixation than the fifth predicted to be fastest (Extended Data Table~\ref{tab:rest_time_all}). The groups that the single map underrepresents can therefore be identified in advance, and they are the youngest and the oldest viewers and Black, lower-income and Republican-leaning viewers.

\begin{figure}[!ht]
\centering
\caption{What the map leaves out.}
\label{fig:outside_map}
\begin{tikzpicture}[/pgfplots/every axis/.append style={scale only axis=true, grid=both, grid style={black!10}, tick label style={font=\footnotesize}, label style={font=\footnotesize}}]
\begin{axis}[name=ta, at={(0cm,0cm)}, anchor=north west, width=4.8cm, height=4.6cm, title={\textbf{a}\,\, Gaze time on text, by age}, title style={font=\small, align=left, at={(0,1)}, anchor=south west}, xtick={1,2,3,4,5,6,7,8,9,10}, xticklabels={{3 to 6},{7 to 9},{10 to 12},{13 to 15},{16 to 18},{19 to 34},{35 to 54},{55 to 64},{65 to 74},{75 and older}}, x tick label style={font=\scriptsize, rotate=40, anchor=north east}, ylabel={percent of fixation time}, ymin=0, ymax=27, xmin=0.4, xmax=10.6]
\addplot[panelblue, thick, mark=*, mark size=1.8pt, error bars/.cd, y dir=both, y explicit, error bar style={panelblue}] coordinates {(1,6.40) -= (0,0.63) += (0,0.71) (2,10.78) -= (0,0.37) += (0,0.37) (3,14.54) -= (0,0.32) += (0,0.37) (4,16.61) -= (0,0.54) += (0,0.54) (5,18.17) -= (0,0.71) += (0,0.72) (6,21.57) -= (0,0.45) += (0,0.44) (7,22.79) -= (0,0.34) += (0,0.40) (8,21.70) -= (0,0.69) += (0,0.74) (9,21.21) -= (0,1.14) += (0,1.06) (10,16.33) -= (0,1.71) += (0,1.82)};
\addplot[panelred, dashed, thick, domain=0.4:10.6] {13.6};
\node[font=\scriptsize, panelred, anchor=north east, inner sep=1pt] at (axis cs:10.5,12.9) {single map, 13.6\%};
\end{axis}
\begin{axis}[name=tb, at={(ta.north east)}, xshift=1.7cm, anchor=north west, width=4.8cm, height=4.6cm, title={\textbf{b}\,\, Viewings reaching text}, title style={font=\small, align=left, at={(0,1)}, anchor=south west}, xtick={1,2,3,4,5,6,7,8,9,10}, xticklabels={{3 to 6},{7 to 9},{10 to 12},{13 to 15},{16 to 18},{19 to 34},{35 to 54},{55 to 64},{65 to 74},{75 and older}}, x tick label style={font=\scriptsize, rotate=40, anchor=north east}, ylabel={percent of viewings}, ymin=0, ymax=72, xmin=0.4, xmax=10.6]
\addplot[panelblue, thick, mark=*, mark size=1.8pt, error bars/.cd, y dir=both, y explicit, error bar style={panelblue}] coordinates {(1,24.52) -= (0,1.83) += (0,1.87) (2,34.43) -= (0,0.83) += (0,0.91) (3,43.24) -= (0,0.81) += (0,0.82) (4,47.56) -= (0,1.21) += (0,1.20) (5,52.46) -= (0,1.62) += (0,1.65) (6,60.73) -= (0,0.90) += (0,0.97) (7,63.46) -= (0,0.79) += (0,0.86) (8,58.16) -= (0,1.58) += (0,1.39) (9,55.73) -= (0,2.32) += (0,2.12) (10,43.43) -= (0,4.01) += (0,4.04)};
\end{axis}
\begin{axis}[name=tc, at={(ta.south west)}, xshift=5.6cm, yshift=-2.7cm, anchor=north west, width=7.4cm, height=6.44cm, title={\textbf{c}\,\, Fixations in one viewing, count}, title style={font=\small, align=left, at={(0,1)}, anchor=south west}, xlabel={first group against second, percent, with a 95\% interval}, ytick={12,11,10,9,8,7,6,5,4,3,2,1}, yticklabels={{75 and older vs 19 to 34, museum, 5.9 and 7.1},{3 to 6 vs 19 to 34, museum, 5.9 and 7.1},{children vs young adults, Wilming and colleagues, 11.9 and 14.6},{older vs young adults, Wilming and colleagues, 15.6 and 14.6},{55 and older vs 18 to 34, news, 10.4 and 11.2},{Black vs White, news, 9.8 and 11.0},{Black vs White, university videos, 2.1/s and 2.3/s},{leans Republican vs Democratic, news on four topics, 9.0 and 11.3},{lower vs upper income, news, 10.4 and 11.2},{Ghana, Turkey vs Germany, Czechia, students, 9.6 and 10.2},{women vs men, news, 10.8 and 10.9},{women vs men, museum adults, 7.0 and 7.2}}, y tick label style={font=\scriptsize, align=right}, ymin=0.4, ymax=12.6, xmin=-34, xmax=20, scaled x ticks=false]
\draw[black, dotted, thick] (axis cs:0,0.4) -- (axis cs:0,12.6);
\addplot[panelblue, only marks, mark=*, mark size=2.2pt, error bars/.cd, x dir=both, x explicit, error bar style={panelblue}] coordinates {(-17.32,12) -= (3.38,0) += (3.54,0) (-17.23,11) -= (2.01,0) += (2.07,0) (-18.64,10) -= (6.76,0) += (7.25,0) (6.98,9) -= (9.96,0) += (10.36,0) (-7.53,8) -= (3.33,0) += (3.05,0) (-10.77,7) -= (3.95,0) += (4.03,0) (-10.65,6) -= (7.09,0) += (8.04,0) (-19.99,5) -= (8.06,0) += (7.74,0) (-6.64,4) -= (2.92,0) += (2.99,0) (-6.11,3) -= (5.17,0) += (5.26,0) (-1.34,2) -= (2.32,0) += (2.47,0) (-2.06,1) -= (1.07,0) += (1.08,0)};
\end{axis}
\end{tikzpicture}
\par\smallskip
{\scriptsize\justifying\textit{Notes.} \textbf{a}, among the 12{,}056 museum visitors, the share of fixation time that falls on the text of the 20 scenes that carry text, by age band, as the mean over viewers of each viewer's mean, with 95\% intervals from 1,000 bootstrap draws of viewers. The dashed line is the share of the four deep networks' pooled map that falls on the same text, 13.6\%, which the single map assigns to every viewer. \textbf{b}, the share of those viewings in which at least one fixation reached the text, by the same bands. \textbf{c}, the number of fixations in one viewing, a count per viewer averaged over the viewer's images, as the difference between two groups in percent of the second group, with the two group means printed in the row label. Image constants are removed, and in the national sample the device constant as well. The university rows are fixations per second on the two GazeBase videos. Older adults in the data of Wilming and colleagues make more fixations than young adults, the one contrast in the panel that runs against the direction of the museum and the national sample.\par}
\end{figure}

\begin{table}[!ht]
\centering
\caption{Biases of a single prediction, measured in the time per fixation.}
\label{tab:rest_time}
\footnotesize
\sbox{\tabbox}{\begin{tabular}{@{}lll rr r c@{}}
\toprule
& & & \multicolumn{2}{c}{Time per fixation, ms} & & \\
\cmidrule(lr){4-5}
Dataset & First group & Second group & First & Second & Difference & 95\% interval \\
\midrule
Museum visitors & 75 and older & 19 to 34 & 396.9 & 323.9 & $+22.5\%$ & $+15.9$ to $+29.2$ \\
Museum visitors & 3 to 6 & 19 to 34 & 383.6 & 323.9 & $+18.5\%$ & $+14.9$ to $+22.1$ \\
Healthy Brain Network & 5 to 7 & 14 to 17 & 277.6 & 238.9 & $+16.2\%$ & $+10.9$ to $+21.8$ \\
National, all news & 55 and older & 18 to 34 & 507.5 & 479.5 & $+5.8\%$ & $+2.3$ to $+9.3$ \\
GazeBase, videos & Black & White & 424.0 & 379.9 & $+11.6\%$ & $+0.7$ to $+22.5$ \\
National, all news & Black & White & 542.8 & 484.8 & $+12.0\%$ & $+7.3$ to $+16.8$ \\
National, four topics & Leans Republican & Leans Democratic & 556.8 & 464.8 & $+19.8\%$ & $+9.1$ to $+30.6$ \\
National, all news & Brackets 1 to 4 & Brackets 9 to 12 & 513.5 & 477.0 & $+7.7\%$ & $+4.7$ to $+11.2$ \\
\bottomrule
\end{tabular}}%
\resizebox{\textwidth}{!}{\usebox{\tabbox}}
\par\smallskip
{\scriptsize\justifying\settabnotew\noindent\parbox{\tabnotew}{\setlength{\emergencystretch}{2em}\textit{Notes.} Mean time per fixation within a viewing, averaged within viewer, the fixation duration the tracker reports in the public datasets and, in the national sample, the interval from the start of one fixation to the start of the next (Methods). The difference is the first group's mean over the second's, minus one, with a 95\% interval from 2,000 bootstrap draws of viewers. National rows remove image and device constants. Extended Data Table~\ref{tab:rest_time_all} gives every contrast with its group sizes.\par}}
\end{table}

\subsection*{A simulation: what a single map costs a slower group}

\begin{figure}[!ht]
\centering
\caption{What a slower group sees of what the single map ranks first.}
\label{fig:feed}
\includegraphics[width=0.92\textwidth]{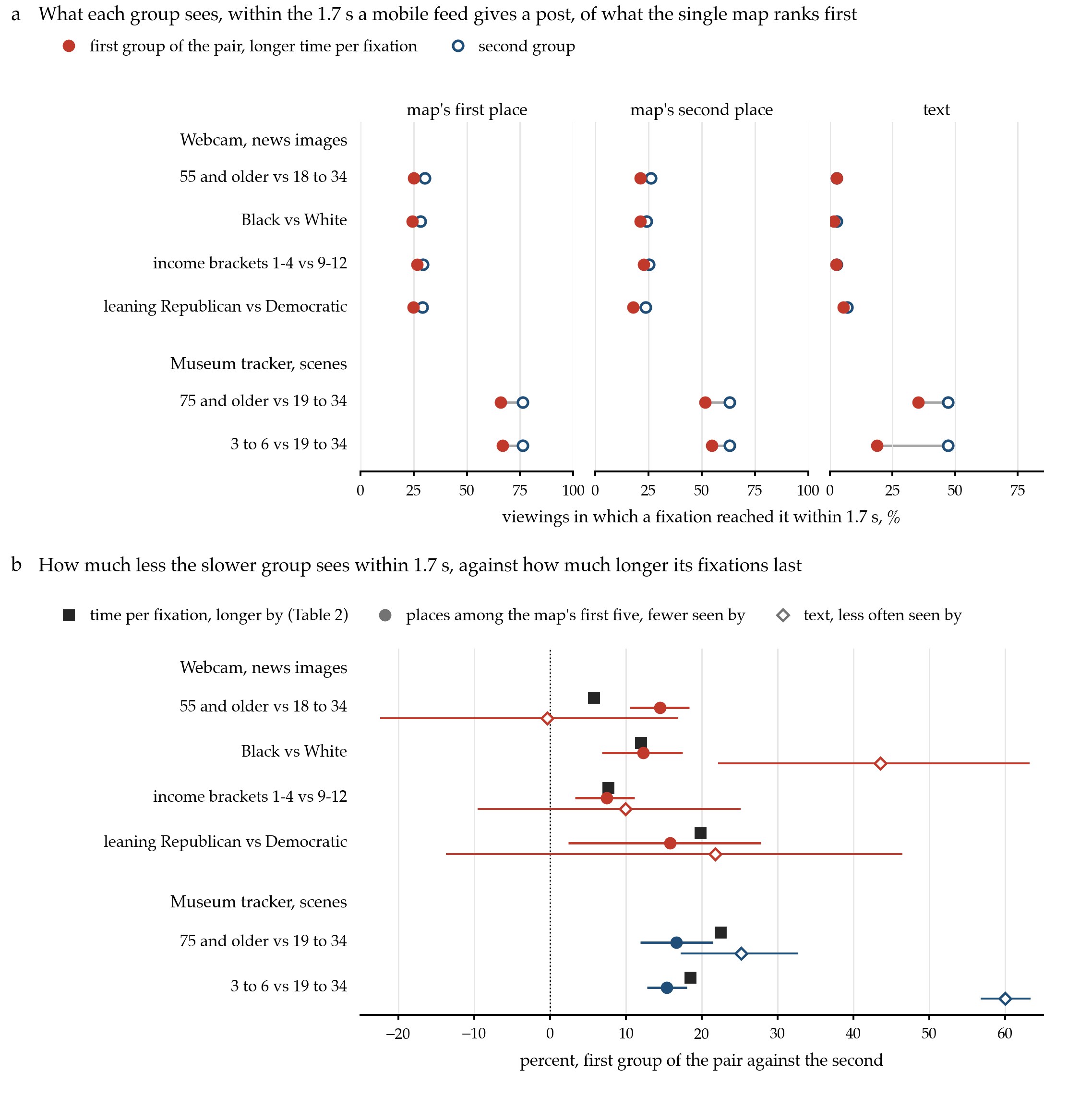}
\par\smallskip
{\scriptsize\justifying\textit{Notes.} \textbf{a}, the share of viewings in which a fixation that began within 1.7~seconds, the time a mobile feed gives a post, reached each of the two places the UNISAL map ranks first on the image, and the share in which it reached a line of text. Filled circles are the first group of each pair, the group with the longer time per fixation in Table~\ref{tab:rest_time}, and open circles the second. The constant of each image and, on the webcam, of the device class are removed. \textbf{b}, squares give how much longer the first group's time per fixation is, circles how many fewer of the map's first five places it reaches within 1.7~seconds, and diamonds how much less often it reaches text, each in percent of the second group, with 95\% intervals from 2,000 bootstrap draws of viewers. Supplementary Information S3 gives the 2.5-second cut, the full viewing, the first fixations whatever time they take, and two other place radii.\par}
\end{figure}

The downstream effect of a biased map is hard to measure directly, since it accumulates over billions of exposures to layouts, crops and feeds that no study can follow. I therefore run a simulation experiment on the recorded viewings, using published exposure times. A post in a social media feed holds the eye for 1.7~seconds on a phone and 2.5~seconds on a desktop \autocite{facebook2016capturing}. I cut every recorded viewing at these times, as a feed would, and record whether a fixation reached the places the single map ranks first, where a layout guided by the map puts a brand or a headline, and whether it reached the text of the image (Fig.~\ref{fig:feed}). Differences in what groups reach matter for layouts designed around a single map, because the groups may not reach the same information within the time available.

In the simulated mobile feed, groups that spend longer at each fixation reach fewer of the locations the single map ranks highest within the same 1.7~seconds (Fig.~\ref{fig:feed}). The shortfall varies across demographic groups. By age, viewers 55 and older reach 14\% fewer of the map's five highest-ranked locations than viewers aged 18 to 34, while museum visitors aged 75 and older are 13\% less likely than young adults to reach the highest-ranked location. Children aged 3 to 6 are 60\% less likely than young adults to reach the text. By race, Black viewers reach 12\% fewer of the five highest-ranked locations than White viewers and are 44\% less likely to reach the text. When the comparison uses the first three fixations regardless of their duration, the shortfall in locations reached falls to 2--10\%, suggesting that differences in fixation duration account for 31--86\% of the original shortfall (Supplementary Information S3).

Published studies link these fixations to what viewers retain. Recognition of a picture grows with the number of fixations made on it \autocite{loftus1972eye}, and memory for an advertised brand follows the fixations its brand receives \autocite{wedel2000eye}. A layout that follows one map therefore gives the slower groups fewer fixations on its message, and every product that uses the map repeats this shortfall for the same groups.

\subsection*{A model of groups}

A fairer model of attention would predict how each group views an image. For a model to predict how a group looks at new images, it must learn patterns of attention that persist across images. The group differences in fixation duration reported in Table~\ref{tab:rest_time} suggest that such patterns exist. I therefore ask whether this habit identifies the group of a viewer held out of training, and whether a model of attention can learn it.

\begin{figure}[!htbp]
\centering
\caption{A model of groups built on how long viewers look.}
\label{fig:model_of_groups}
\includegraphics[width=0.92\textwidth]{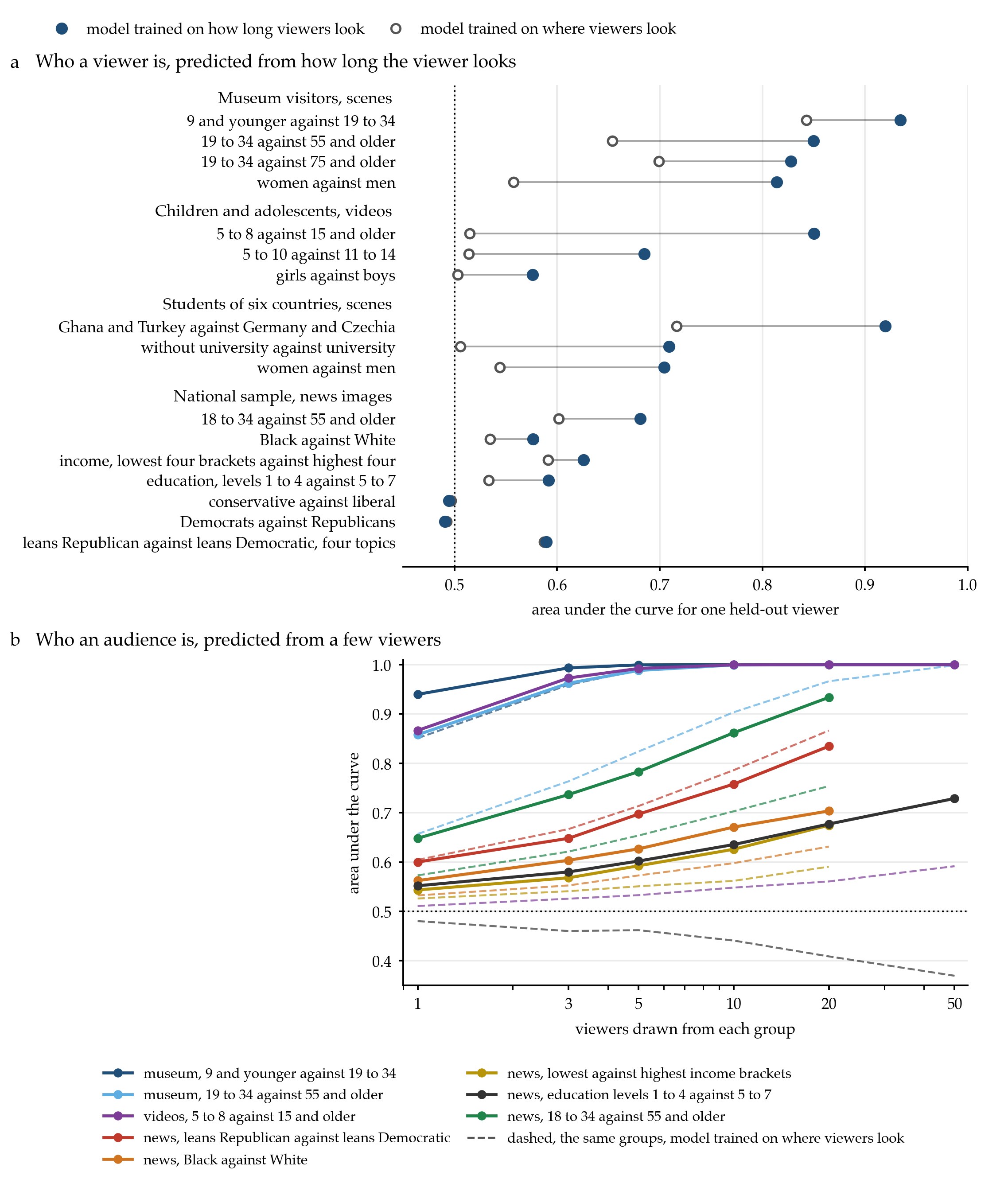}
\par\smallskip
{\scriptsize\justifying\textit{Notes.} \textbf{a}, the area under the curve with which a gradient-boosting classifier predicts the group of a viewer held out of its training. The filled circles are the model trained on how long the viewer looks and the open circles the model trained on where the viewer looks, with the features of each in Methods. Both run in five folds by viewer over three repetitions, with every reference built from the training viewers of the fold. The national sample enters with the recruitment wave and the device class as the unit within which every reference is taken, and with the raw webcam stream added to the fixations. \textbf{b}, audience identification. Each viewer's held-out probability, the mean of the three repetitions, is averaged over $n$ viewers drawn at random from each group, in the national sample from the same recruitment wave with the area under the curve computed within each wave and averaged over the waves, and the area under the curve between the two sets of means is drawn against $n$ over 4,000 draws per size, for the duration model (solid) and the place model (dashed). Extended Data Table~\ref{tab:group_model} gives every axis.\par}
\end{figure}

Two gradient-boosting classifiers answer the first question (Methods, Extended Data Table~\ref{tab:group_model}). One sees how a viewer divides the viewing time, and the other sees only where the viewer looks, the information a saliency map could learn. Accuracy is again the AUC (Methods). Viewing time identifies the group far better than place. On museum trackers, the duration-based classifier distinguishes visitors 55 and older from young adults with an AUC of 0.85, compared with 0.65 for the location-based classifier. Among children in the Healthy Brain Network, the corresponding AUCs are 0.85 and 0.51 for distinguishing ages 5 to 8 from 15 and older, with standard errors of 0.01 to 0.02 (Fig.~\ref{fig:model_of_groups}a). On the webcam, a single viewer carries less information, but averaging predictions over 10 viewers drawn from the same recruitment wave identifies an audience by age with an AUC of 0.86 (95\% interval 0.79 to 0.91) and by race with 0.67 (0.57 to 0.74), while party identification remains at chance, 0.47 (0.39 to 0.54; Fig.~\ref{fig:model_of_groups}b).

A neural network that predicts each next fixation and its duration once it is told the viewer’s group answers the second question (Methods, Extended Data Table~\ref{tab:group_network}). On museum visitors held out of its training, it distinguishes children from young adults with an AUC of 0.83 and visitors 55 and older from young adults with an AUC of 0.68, above all 20 copies of the network trained on shuffled group labels, which reach at most 0.73 and 0.65, respectively. For children the place of the next fixation carries most of the difference (0.83 against 0.75 for its duration), and for visitors 55 and older the duration does (0.70 against 0.63 for the place). On the webcam the same network gives 0.50 for age and 0.57 for race, below its best shuffled copies at 0.59 and 0.64. A second network, which reads the raw camera stream, beats all its shuffled copies for age (0.61 against 0.59) and race (0.61 against 0.60), and also for four of the 20 national contrasts beyond the four chosen in advance, among them sex (0.54 against 0.51), always by 0.004 to 0.023 (Extended Data Table~\ref{tab:group_network}). A model of attention therefore learns how groups view on the museum tracker and learns it only weakly on the webcam.

Training models on human attention should reflect what the eye-tracking data reveal about the sources of variation in viewing. Since groups largely agree on where to look, models can learn the spatial distribution of attention from pooled gaze while using viewer characteristics to learn differences in fixation duration. The results also show why the choice of gaze signal matters. The network that reads the raw gaze stream detects some group differences on the webcam that the network on fixations misses, by small margins. This suggests that training should preserve the temporal information available in each recording.

\section*{Discussion}

The best computer vision models predict where people look almost as well as other people do, yet this accuracy conceals systematic differences across sociodemographic and political groups. Although viewers tend to fixate on the same locations, they differ in how long they look at each one, which affects how much of an image they explore. Models trained on pooled gaze overlook these differences, which means they reproduce some groups' visual attention patterns more accurately than others and make those groups' patterns the default against which visual AI systems learn to interpret the world.

This bias carries over into a number of applications that use saliency predictions.  Saliency maps guide image cropping, advertising layouts, website design, the placement of information on screens and many others. A difference of a few fixations may seem inconsequential, but the same model can shape billions of encounters with visual information. In the feed simulation, groups with longer fixations consistently reach fewer of the locations the model ranks highest. Whether these differences accumulate into unequal memory or political knowledge is not yet known. What is known is that the same prediction can repeatedly disadvantage the same audiences, without anyone designing the downstream application to discriminate.

This study also shows how to address this bias. I identify gaze timing as a criterion for distinguishing visual attention patterns across sociodemographic and political groups and demonstrate how models can learn these differences from viewer characteristics and fixation durations (Extended Data Tables~\ref{tab:group_model} and~\ref{tab:group_network}). This provides a basis for training models that simulate visual attention for specific audiences, which could help researchers and designers adapt visual communication to the ways different people actually look at images. Such models, all else equal, could help make visual information accessible to people with different attention patterns by accounting for differences in how people view the same content.

\section*{Methods}
\label{sec:methods}

\subsection*{Data}
The analyses use nine eye-tracking datasets, which record the gaze position of each viewer over time (Supplementary Table~\ref{tab:datasets}). Eight record who their viewers are, the national sample of this study and seven public datasets. The ninth, MIT1003, is the laboratory set that trained DeepGazeIIE and TranSalNet. Together they hold 27{,}399 viewers and about 10~million fixations, the intervals between saccades, the rapid jumps of the eye, in which the gaze holds one location.

\subsection*{National sample}
In total 2,993 US adults took part, giving 3,023 records. They were recruited in four waves to match population shares by age, gender, education, race, income and party. The sample has 58.6\% women, 45.8\% Democrats and 38.4\% Republicans (Extended Data Table~\ref{tab:sample_composition}). Each person viewed 16 to 28 news images from Getty Images, each for 5 seconds. Waves 1 to 3 drew from 35 images on immigration and wave~4 from 48 images on immigration, gun control, the January 6 attack and LGBTQ+ rights. Position on the issues is the mean of a viewer's positions on the four topics of wave~4, each coded from $-2$ to $+2$ toward the Democratic side, and in waves 1 to 3 it is the position on immigration, the one topic of those waves. The party thermometers are the 0 to 100 ratings of warmth toward each party.  A viewing is one person looking at one image, and the sample holds 61{,}458 viewings. 

\subsection*{Public corpora}
Linka and colleagues recorded 12{,}056 museum visitors viewing everyday scenes \autocite{linka2025protracted}. \v{C}en\v{e}k and colleagues recorded 408 students in seven countries, the Taiwan sample on a different tracker from the rest \autocite{cenek2025cross}. Wilming and colleagues recorded 58 people in three age groups \autocite{wilming2017extensive}. Strauch and colleagues recorded museum visitors viewing a single image \autocite{strauch2023saliency}. FakeNewsPerception recorded 25 people reading news pages \autocite{sumer2021fakenewsperception}. The Healthy Brain Network recorded children and teenagers watching videos \autocite{alexander2017open}, and GazeBase recorded 322 students watching videos \autocite{griffith2021gazebase}. MIT1003 recorded 15 people in a laboratory, and I use 250 of its scenes \autocite{judd2009learning}. Extended Data Table~\ref{tab:sample_composition}b describes each dataset. Six are scored against the models (Extended Data Table~\ref{tab:public_models}), and five give fixation durations (Extended Data Table~\ref{tab:rest_time_all}).

\subsection*{Ethics}
The Hertie School Ethics Committee approved the national sample (application 20230518-46), and the University of North Carolina at Chapel Hill (23-1048) and Florida State University (STUDY00004983, STUDY00006310) exempted it from full review. Participants consented to the study and, separately, to the eye tracking. The public datasets enter as their authors released them, without names, under the consent their publications describe.

\subsection*{Gaze signals in the national sample}
Gaze was recorded through each participant's own webcam with WebGazer, an open-source JavaScript library that estimates from the webcam image where on the screen a person looks about 40 times a second \autocite{papoutsaki2016webgazer, semmelmann2018online, yang2021webcam, crowdsourced2025quality}. This series of estimates, the raw stream, is the main signal. Estimates that fall off the image, 16.4\% of the stream, are dropped, which leaves 11.4 million, and every participant who finished the study stays in. Recording quality enters through the share of gaze on the screen that the survey platform exports and related measures of each viewing. Fixations, runs of estimates that stay within 3\% of the frame for at least 80~ms, give the time per fixation, the number of fixations and a second version of every score. The webcam's error is measured for each viewer as the distance between where the viewer looked on average and where everyone else looked on the same image. Its median is 3.72 degrees of visual angle against 0.76 in the laboratory, for an image 30 degrees wide (Extended Data Fig.~\ref{fig:measurement_error}).

\subsection*{Saliency models and scores}
A saliency model computes from an image a saliency map, which assigns each location a value that ranks how likely viewers are to fixate it. I test four deep networks trained on data of human attention, DeepGazeIIE \autocite{linardos2021deepgaze}, TranSalNet-Dense and TranSalNet-Res \autocite{lou2022transalnet} and UNISAL \autocite{droste2020unisal}, and two classical models that derive saliency from low-level image features, Spectral Residual \autocite{hou2007saliency} and Fine-Grained \autocite{montabone2010human}.

The score is the area under the receiver operating characteristic curve (AUC), the share of pairs in which a map ranks a place a person looked at above a comparison place, so 0.5 is chance and 1 is perfect. Each score uses 2,000 comparison places per image. The raw score draws comparison places anywhere in the image. People look near the middle of almost any picture, so the center-corrected score draws comparison places from where people looked on other images, which cancels this shared habit \autocite{zhang2008sun, bylinskii2019different}. The baseline is the center map, a Gaussian centered on the image that ignores its content, with a width of 0.25~of the image. A model's margin is its score minus the center map's score on the same gaze. Some tables divide each map by the model's average map, which removes what the model draws on every image \autocite{kummerer2018saliency}. Extended Data Table~\ref{tab:saliency_predictors} relates each map to closeness to the center, people, contrast and object size.

\subsection*{Reference panel of viewers}
To learn how well people themselves predict a new viewer, a map built from other viewers of the same image is scored on viewers set aside, for panels of 1 to 256 people (Extended Data Table~\ref{tab:center_increment}b).

\subsection*{Group readouts and own pools}
I ask whether a map built for one group predicts that group better than a map built for everyone. A readout, a small neural network that predicts a group's fixations from the features of a pretrained image network and from the released maps, learns from each group and from a mixed group of the same size. The own pool asks the same with a density map of the gaze of half of a group's viewers. The test of every readout and every pool uses only images and viewers outside its training data (Extended Data Table~\ref{tab:group_readouts}, Fig.~\ref{fig:map_agree}). Standard errors and 95\% intervals come from 1,000 bootstrap draws, each of which resamples the images and, within each group, the viewers together with all their images.

\subsection*{Number of fixations, text and the feed experiment}
The number of fixations and the share of time spent on text, which optical character recognition locates in each image, are compared between groups after removing what is typical for each image and device (Extended Data Table~\ref{tab:places_text}). The feed experiment cuts each viewing at 1.7 and 2.5 seconds, the time a post holds the eye in a feed on a phone and on a computer \autocite{facebook2016capturing}, and counts how many of the map's top five places and how much text each group reaches.

\subsection*{Model of groups}
A gradient-boosting classifier, an ensemble of decision trees in which each new tree corrects the errors of the trees before it, learns to tell a person's group from how long the person's fixations last and where they land. The duration model sees the number of fixations and their durations, how the viewer divides fixation time among the categories of the image, how much longer the viewer stays than training viewers at the same places, and how closely the viewer's time matches each group's time map; in the national sample it also sees the speed of the raw stream. The place model sees where the fixations land, how far they spread, which categories they fall on and how closely they match each group's place map. It is trained on some people and tested on others it never saw, in four datasets (Extended Data Table~\ref{tab:group_model}). Accuracy is again the AUC, here the probability that a randomly drawn member of the first group receives a higher predicted probability than a randomly drawn member of the second, with 0.5 as chance. Checks on the recording conditions are in Supplementary Information S4.

\subsection*{Network model of groups}
A neural network predicts where a person looks next and for how long, once told the person's group (Supplementary Information S5). It learns from four datasets, 533{,}836 viewings of 442 images. Before training, the people were split into five parts. The network was built on three parts, a rule fixed in advance chose between two ways of training on the fourth, and the final network was scored once on the fifth (Extended Data Table~\ref{tab:group_network}). A group contrast counts as learned when it beats 20 networks trained with shuffled group labels.

\subsection*{Time per fixation}
The time per fixation is a person's average fixation length, compared between groups as a percentage after removing what is typical for each image and device (Table~\ref{tab:rest_time}, Extended Data Table~\ref{tab:rest_time_all}). On the webcam a fixation lasts until the next one starts, so the time per fixation and the number of fixations measure one difference. Timed from its first estimate to its last, a webcam fixation lasts 122~ms on average, and on that measure age and race differ by $-1.7$\% [$-3.0$, $-0.4$] and $-0.9$\% [$-2.3$, $+0.5$]. When I hold the quality of the recording fixed, the age gap is $+3.9$\% [$+0.9$, $+6.7$] and the race gap $+7.7$\% [$+3.7$, $+11.7$]. Adding the gaze-to-click error, the distance between where viewers clicked and where the webcam placed their gaze, which the study recorded for 319 viewers, gives $+5.5$\% [$+2.5$, $+8.3$] for age and $+7.4$\% [$+3.5$, $+11.3$] for race. On computers in waves 2 to 4, which recorded no phones, the income gap is $+6.1$\% [$+1.3$, $+10.8$]. On the four topics the gap by position is $+19.6$\% [$+9.4$, $+29.3$] when I also hold age and race fixed and $+17.4$\% [$+8.1$, $+27.5$] when I add gender, education and income.

\subsection*{Inference}
Group gaps in the national sample come from regressions with fixed effects, a separate intercept for each image and for each combination of wave and device. Standard errors cluster by participant, so a person's viewings may correlate with each other. Extended Data Table~\ref{tab:spec_grid} repeats each gap under alternative choices and reports UNISAL. No result is corrected for multiple comparisons (Supplementary Information S1).

\subsection*{Data availability}
The gaze data that support the findings of this study, comprising the raw gaze streams, the extracted fixations, and the participant attributes used in every analysis, are deposited in a public repository and released openly on publication, with access for the editors and the referees during review. The Getty Images images used as stimuli are licensed, so the stimulus identifiers are released and the rights holder supplies the files. MIT1003 is publicly available from its original distribution \autocite{judd2009learning}. The public datasets are available from the repositories named in their publications \autocite{linka2025protracted, cenek2025cross, wilming2017extensive, strauch2023saliency, sumer2021fakenewsperception, alexander2017open, griffith2021gazebase}.

\subsection*{Code availability}
Every script that produces a number in this paper is deposited in the same repository, released on publication and available to the editors and referees during review.

\subsection*{Author contributions}
The author designed the study and performed the analyses. The author wrote the paper.

\subsection*{Competing interests}
The author declares no competing interests.

\subsection*{Additional information}
Supplementary Information is available for this paper. Correspondence and requests for materials should be addressed to the author. Reprints and permissions information is available at www.nature.com/reprints.

\begingroup\raggedright\sloppy
\printbibliography
\endgroup

\newpage
\appendix
\setcounter{table}{0}
\setcounter{figure}{0}
\renewcommand{\thetable}{\arabic{table}}
\renewcommand{\thefigure}{\arabic{figure}}
\renewcommand{\tablename}{Extended Data Table}
\renewcommand{\figurename}{Extended Data Fig.}

\section*{Extended Data}

\begin{table}[H]
\centering
\caption{Sample composition of every dataset.}
\label{tab:sample_composition}
\small
\renewcommand{\arraystretch}{1.15}
\textbf{a.} National sample of this study\par\smallskip
\sbox{\tabbox}{\begin{tabular}{@{}l l r r@{}}
\toprule
Characteristic & Category & $n$ & \% \\
\midrule
\textbf{Age} & 18~to~34 & 888 & 29.4 \\
 & 35~to~54 & 1{,}368 & 45.3 \\
 & 55~to~74 & 715 & 23.7 \\
 & 75 or older & 52 & 1.7 \\
\addlinespace
\textbf{Gender} & Men & 1{,}252 & 41.4 \\
 & Women & 1{,}771 & 58.6 \\
\addlinespace
\textbf{Race} & White & 2{,}308 & 76.3 \\
 & Black or African American & 336 & 11.1 \\
 & Asian & 161 & 5.3 \\
 & Other or multiple races & 216 & 7.1 \\
\addlinespace
\textbf{Education} & High school or less & 511 & 16.9 \\
 & Some college or associate & 1{,}059 & 35.0 \\
 & Bachelor's degree & 980 & 32.4 \\
 & Graduate degree & 472 & 15.6 \\
\addlinespace
\textbf{Household income} & Lower (codes 1~to~4) & 914 & 30.2 \\
 & Middle (codes 5~to~8) & 1{,}037 & 34.3 \\
 & Upper (codes 9~to~12) & 1{,}071 & 35.4 \\
\addlinespace
\textbf{Party ID} & Democrat & 1{,}386 & 45.8 \\
 & Independent & 425 & 14.1 \\
 & Republican & 1{,}162 & 38.4 \\
\addlinespace
\textbf{Partisan strength} & Strong partisan & 1{,}031 & 34.1 \\
 & Not strong or leaning & 1{,}517 & 50.2 \\
 & Independent & 425 & 14.1 \\
\addlinespace
\textbf{Ideology} & Conservative & 1{,}037 & 34.3 \\
 & Moderate & 817 & 27.0 \\
 & Liberal & 1{,}169 & 38.7 \\
\bottomrule
\end{tabular}}%
\usebox{\tabbox}
\par\smallskip
{\scriptsize\justifying\settabnotew\noindent\parbox{\tabnotew}{\setlength{\emergencystretch}{2em}\textit{Notes.} The table counts 3{,}023 records of 2{,}993 participants, since some people took part twice. Every percentage is a share of all 3{,}023 records, so a characteristic that some participants left unanswered sums to slightly under 100\%. In the data file the age codes of wave~4 sit one step below those of the other waves, so they were raised by one step. Education is asked on a 7-point scale. Income is asked in 12 brackets, from under \$10k to over \$150k. Ideology is a 7-point scale on which 1 is very conservative and 7 very liberal.\par}}
\end{table}

\begin{table}[H]
\ContinuedFloat
\centering
\caption{Sample composition of every dataset, continued.}
\scriptsize
\renewcommand{\arraystretch}{0.9}
\textbf{b.} Public datasets\par\smallskip
\noindent\begin{minipage}[t]{0.49\textwidth}\centering
\begin{tabular}[t]{@{}l l r r@{}}
\toprule
Characteristic & Category & $n$ & \% \\
\midrule
\multicolumn{4}{@{}l}{\textit{Museum visitors, Linka et al.}, 12{,}056 viewers} \\
Age & 3 to 9 & 2{,}175 & 18.0 \\
 & 10 to 18 & 4{,}884 & 40.5 \\
 & 19 to 34 & 1{,}825 & 15.1 \\
 & 35 to 54 & 2{,}087 & 17.3 \\
 & 55 to 74 & 784 & 6.5 \\
 & 75 to 95 & 126 & 1.0 \\
 & unknown & 175 & 1.5 \\
Gender & women & 5{,}137 & 42.6 \\
 & men & 6{,}368 & 52.8 \\
 & other & 551 & 4.6 \\
Language & German & 11{,}210 & 93.0 \\
 & English & 846 & 7.0 \\
Glasses & without glasses & 8{,}603 & 71.4 \\
 & with glasses & 3{,}453 & 28.6 \\
\addlinespace[2pt]
\multicolumn{4}{@{}l}{\textit{Students of six countries, \v{C}en\v{e}k et al.}, 408 viewers} \\
Sample & Czechia & 92 & 22.5 \\
 & Taiwan & 65 & 15.9 \\
 & Turkey, east & 57 & 14.0 \\
 & Bulgaria & 56 & 13.7 \\
 & Germany & 54 & 13.2 \\
 & Ghana & 42 & 10.3 \\
 & Turkey, west & 42 & 10.3 \\
Sex & women & 216 & 52.9 \\
 & men & 190 & 46.6 \\
 & other & 2 & 0.5 \\
Education & university & 250 & 61.3 \\
 & secondary school & 126 & 30.9 \\
 & vocational school & 32 & 7.8 \\
Family class & poverty & 3 & 0.7 \\
 & lower class & 46 & 11.3 \\
 & middle class & 275 & 67.4 \\
 & upper middle class & 77 & 18.9 \\
 & upper class & 7 & 1.7 \\
Age & 21 or younger & 177 & 43.4 \\
 & 22 & 81 & 19.9 \\
 & 23 or older & 150 & 36.8 \\
\addlinespace[2pt]
\multicolumn{4}{@{}l}{\textit{Wilming et al.}, 58 viewers} \\
Age group & children (mean age 7.6) & 18 & 31.0 \\
 & young adults (22.1) & 23 & 39.7 \\
 & older adults (80.6) & 17 & 29.3 \\
\addlinespace[2pt]
\addlinespace[2pt]
\multicolumn{4}{@{}l}{\textit{MIT1003, Judd et al.}, 15 viewers} \\
\multicolumn{4}{@{}l}{viewer attributes unrecorded} \\
\bottomrule
\end{tabular}
\end{minipage}\hfill\begin{minipage}[t]{0.49\textwidth}\centering
\begin{tabular}[t]{@{}l l r r@{}}
\toprule
Characteristic & Category & $n$ & \% \\
\midrule
\multicolumn{4}{@{}l}{\textit{Museum, one image, Strauch et al., Tobii Pro Fusion}, 1{,}233 viewers} \\
Age & under 18 & 284 & 23.0 \\
 & 18 to 34 & 564 & 45.7 \\
 & 35 to 54 & 297 & 24.1 \\
 & 55 or older & 29 & 2.4 \\
 & unknown & 59 & 4.8 \\
Gender & men & 640 & 51.9 \\
 & women & 499 & 40.5 \\
 & other & 94 & 7.6 \\
\addlinespace[2pt]
\multicolumn{4}{@{}l}{\textit{Museum, one image, Strauch et al., Tobii 4C}, 6{,}909 viewers} \\
Age & under 18 & 854 & 12.4 \\
 & 18 to 34 & 2{,}801 & 40.5 \\
 & 35 to 54 & 1{,}553 & 22.5 \\
 & 55 or older & 243 & 3.5 \\
 & default year 2000, or unknown & 1{,}458 & 21.1 \\
Gender & men & 3{,}519 & 50.9 \\
 & women & 2{,}906 & 42.1 \\
 & other & 484 & 7.0 \\
\addlinespace[2pt]
\multicolumn{4}{@{}l}{\textit{News pages, S\"umer et al.}, 25 viewers} \\
Age & 26 or younger & 14 & 56.0 \\
 & older than 26 & 11 & 44.0 \\
Gender & women & 13 & 52.0 \\
 & men & 12 & 48.0 \\
Education & Bachelor & 8 & 32.0 \\
 & Abitur & 8 & 32.0 \\
 & Master & 7 & 28.0 \\
 & state examination & 2 & 8.0 \\
Politics & 2 to 3 & 11 & 44.0 \\
 & 4 & 6 & 24.0 \\
 & 5 to 8 & 8 & 32.0 \\
\addlinespace[2pt]
\multicolumn{4}{@{}l}{\textit{Young viewers, Healthy Brain Network}, 3{,}380 viewers} \\
Age & 5 to 8 & 983 & 29.1 \\
 & 9 to 11 & 728 & 21.5 \\
 & 12 to 14 & 413 & 12.2 \\
 & 15 or older & 302 & 8.9 \\
 & unknown & 954 & 28.2 \\
Sex & girls & 865 & 25.6 \\
 & boys & 1{,}561 & 46.2 \\
 & unknown & 954 & 28.2 \\
\addlinespace[2pt]
\multicolumn{4}{@{}l}{\textit{University students, GazeBase}, 322 viewers} \\
Ethnicity & White & 177 & 55.0 \\
 & Black & 29 & 9.0 \\
 & Hispanic & 80 & 24.8 \\
 & Asian & 10 & 3.1 \\
 & other or multiple & 26 & 8.1 \\
Gender & men & 171 & 53.1 \\
 & women & 151 & 46.9 \\
Age & 21 or younger & 210 & 65.2 \\
 & older than 21 & 112 & 34.8 \\
\bottomrule
\end{tabular}
\end{minipage}
\par\smallskip
{\scriptsize\justifying\noindent\parbox{\textwidth}{\setlength{\emergencystretch}{2em}\textit{Notes.} A viewer is counted when the dataset's gaze file, as it enters the analyses, holds fixations for that person, and the characteristics come from the participant files the authors released. Every percentage is a share of all viewers of the dataset. Museum ages below 3 and above 95 are treated as unknown. On the Tobii 4C, age is the year of the visit minus the year of birth, and 1{,}458 visitors kept the year 2000 that the entry form showed by default, so their age is unknown. The Healthy Brain Network count includes the viewers with fixations from both eyes, and 954 of them lack age and sex in the open release. The mean ages of the three groups of Wilming and colleagues come from the original publication. GazeBase ethnicity is self-described and grouped by its wording, and its age is split at the median, as is the age of the news-page readers. The politics of the news-page readers is grouped on the scale the authors released.\par}}
\end{table}

\begin{table}[H]
\centering
\caption{Deep models follow the center, classical ones the contrast.}
\label{tab:saliency_predictors}
{\footnotesize\textbf{a. The four predictors on the 83 news images}}\par\smallskip
\begingroup
\small
\sbox{\tabbox}{\begin{tabular}{llrrrrr}
\toprule
Model & Family & Center & Persons & Contrast & Size & $R^2$ \\
\midrule
DeepGazeIIE       & deep      & \textbf{0.425} & 0.048 & 0.083 & $-0.017$ & 0.208 \\
TranSalNet-Dense  & deep      & \textbf{0.550} & 0.072 & 0.106 & $0.036$ & 0.380 \\
TranSalNet-Res    & deep      & \textbf{0.554} & 0.056 & 0.102 & $0.035$ & 0.374 \\
UNISAL            & deep      & \textbf{0.467} & 0.139 & 0.118 & $-0.007$ & 0.308 \\
Spectral Residual & classical & 0.097 & 0.027 & \textbf{0.386} & $-0.003$ & 0.171 \\
Fine-Grained      & classical & 0.063 & 0.032 & \textbf{0.536} & $0.004$ & 0.305 \\
\bottomrule
\end{tabular}}%
\usebox{\tabbox}
\par\smallskip
{\scriptsize\justifying\settabnotew\noindent\parbox{\tabnotew}{\setlength{\emergencystretch}{2em}\textit{Notes.} Each row regresses a model's map on four properties of each place in the image, over all 83 news images together, with every variable standardized within each image. A coefficient therefore says by how many standard deviations the map changes when the property rises by one standard deviation, and the largest coefficient of each row is in bold. $R^2$ is the share of the variation in the map that the four properties explain together. Center is closeness to the center of the frame, measured as minus the distance from it. Persons marks a place where a person was detected. Contrast is the local difference in brightness. Size is the area of the largest detected object covering the place.\par}}
\endgroup
\par\medskip
{\footnotesize\textbf{b. The models behave the same on news images as on laboratory images}}\par\smallskip
\begingroup
\small
\sbox{\tabbox}{\begin{tabular}{@{}l rr rr rr@{}}
\toprule
 & \multicolumn{2}{c}{Center} & \multicolumn{2}{c}{Contrast} & \multicolumn{2}{c}{$R^2$} \\
\cmidrule(lr){2-3}\cmidrule(lr){4-5}\cmidrule(lr){6-7}
Model & News & MIT1003 & News & MIT1003 & News & MIT1003 \\
\midrule
DeepGazeIIE       & 0.425 & 0.445 & 0.083 & 0.121 & 0.208 & 0.248 \\
TranSalNet-Dense  & 0.550 & 0.532 & 0.106 & 0.144 & 0.380 & 0.382 \\
TranSalNet-Res    & 0.554 & 0.527 & 0.102 & 0.148 & 0.374 & 0.376 \\
UNISAL            & 0.467 & 0.493 & 0.118 & 0.148 & 0.308 & 0.335 \\
Spectral Residual & 0.097 & 0.047 & 0.386 & 0.473 & 0.171 & 0.249 \\
Fine-Grained      & 0.063 & 0.069 & 0.536 & 0.667 & 0.305 & 0.471 \\
\bottomrule
\end{tabular}}%
\usebox{\tabbox}
\par\smallskip
{\scriptsize\justifying\settabnotew\noindent\parbox{\tabnotew}{\setlength{\emergencystretch}{2em}\textit{Notes.} Panel a repeated on the 250 laboratory images of MIT1003, the benchmark that trained DeepGazeIIE and TranSalNet. On both sets the deep networks follow closeness to the center most strongly, and the classical detectors follow sharp changes in brightness. Between the two sets each number moves by at most 0.05 for the center and 0.13 for contrast. The models therefore weigh the same image properties on news and laboratory images.\par}}
\endgroup
\end{table}

\begin{figure}[H]
\centering
\caption{How a map earns a score.}
\label{fig:score_pipeline}
\includegraphics[width=\textwidth]{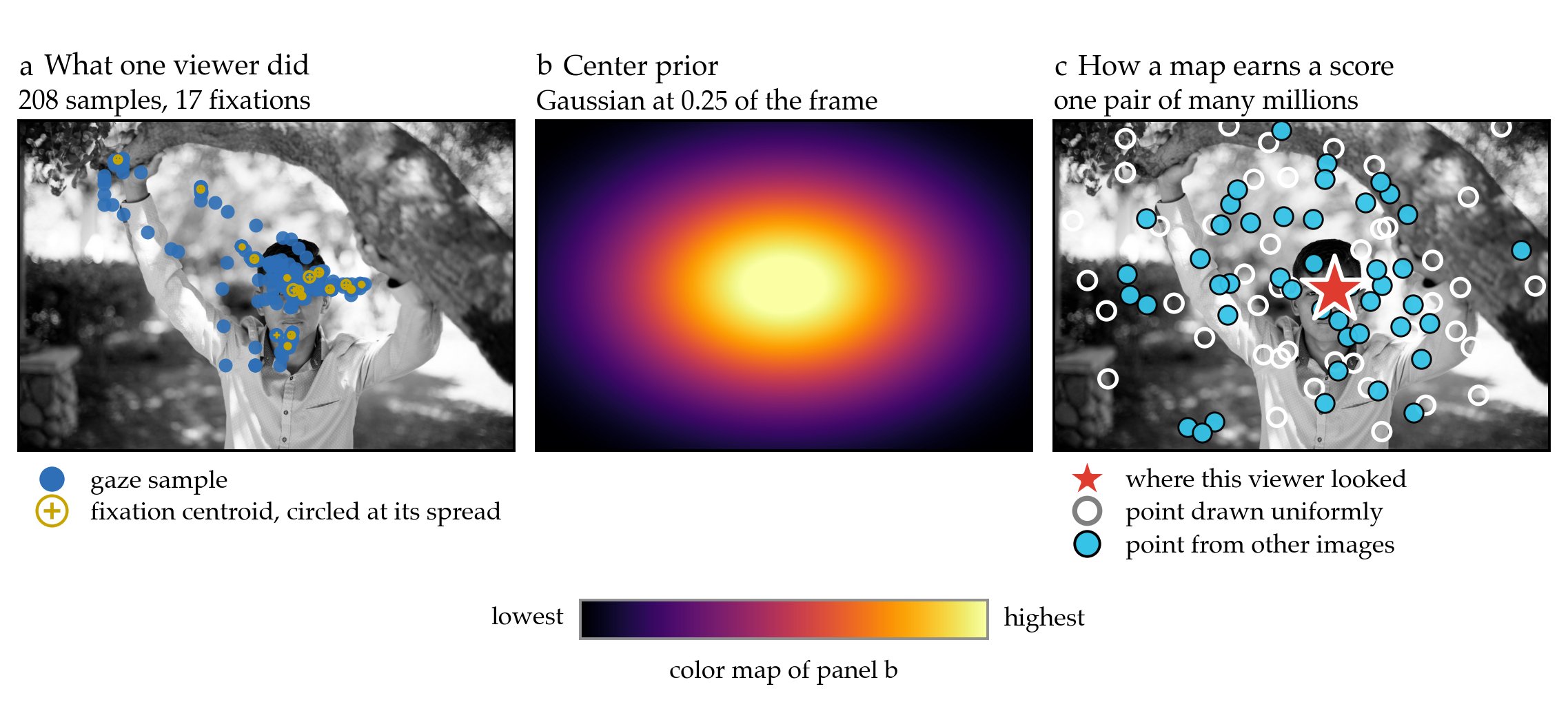}
\par\smallskip
{\scriptsize\justifying\textit{Notes.} The image is the one of the 83 news images with the most viewers. \textbf{a}, a typical viewer of it, the viewer with the median number of gaze estimates among viewings with at least 10 fixations. The dots are the 208~estimates the webcam produced over~5~seconds, and the 17~fixations extracted from them are drawn at their centers and circled at the spread of their estimates. \textbf{b}, the center map, a Gaussian bright spot in the middle of the frame that ignores the content of the image. \textbf{c}, the comparison behind every score. The star marks the fixation of panel a that gathered the most estimates, and a score asks whether the map values the star's place above a comparison place. The open circles are comparison places drawn uniformly in the frame, used by the raw score, and the filled circles are comparison places drawn from gaze on other images, used by the center-corrected score. The panel shows 45 of each kind, a random subset of the 2,000 of each kind that the score of this image uses.\par}
\end{figure}

\begin{figure}[H]
\centering
\caption{The gap holds across the viewer's measurement error.}
\label{fig:measurement_error}
\begin{tikzpicture}
\begin{groupplot}[group style={group size=2 by 1, horizontal sep=3.0cm}, width=0.35\textwidth, height=6.4cm,
  grid=both, grid style={black!10}, title style={font=\small, at={(0,1.02)}, anchor=south west, align=left},
  tick label style={font=\small}, label style={font=\small}, scaled y ticks=false,
  yticklabel style={/pgf/number format/fixed, /pgf/number format/fixed zerofill, /pgf/number format/precision=2},
  legend cell align=left, legend style={font=\small, draw=none, fill=none}]
\nextgroupplot[title={\textbf{a}\,\, Webcam viewers sorted\\by recording error}, xlabel={measured per-viewer error, degrees},
  ylabel={model minus center map}, xmin=0.28, xmax=9.03, ymin=-0.215, ymax=0.008,
  legend style={font=\footnotesize, draw=none, fill=none, at={(0.5,-0.30)}, anchor=north, legend columns=2}]
\addplot[panelblue, thick, mark=*, mark size=1.4pt] coordinates {(0.836,-0.0280) (1.570,-0.0301) (2.150,-0.0277) (2.714,-0.0294) (3.293,-0.0325) (3.936,-0.0331) (4.684,-0.0301) (5.478,-0.0323) (6.652,-0.0246) (8.826,-0.0103)};
\addlegendentry{DeepGazeIIE}
\addplot[panelgreen, thick, mark=square*, mark size=1.4pt] coordinates {(0.836,-0.0423) (1.570,-0.0432) (2.150,-0.0380) (2.714,-0.0381) (3.293,-0.0422) (3.936,-0.0416) (4.684,-0.0392) (5.478,-0.0382) (6.652,-0.0291) (8.826,-0.0117)};
\addlegendentry{TranSalNet-Dense}
\addplot[panelpurple, thick, mark=triangle*, mark size=1.6pt] coordinates {(0.836,-0.0385) (1.570,-0.0396) (2.150,-0.0354) (2.714,-0.0356) (3.293,-0.0396) (3.936,-0.0374) (4.684,-0.0370) (5.478,-0.0363) (6.652,-0.0280) (8.826,-0.0129)};
\addlegendentry{TranSalNet-Res}
\addplot[panelorange, thick, mark=diamond*, mark size=1.8pt] coordinates {(0.836,-0.0244) (1.570,-0.0298) (2.150,-0.0312) (2.714,-0.0260) (3.293,-0.0355) (3.936,-0.0343) (4.684,-0.0344) (5.478,-0.0375) (6.652,-0.0305) (8.826,-0.0175)};
\addlegendentry{UNISAL}
\addplot[black!55, thick, mark=o, mark size=1.4pt] coordinates {(0.836,-0.1851) (1.570,-0.1816) (2.150,-0.1727) (2.714,-0.1638) (3.293,-0.1630) (3.936,-0.1546) (4.684,-0.1392) (5.478,-0.1262) (6.652,-0.0903) (8.826,-0.0019)};
\addlegendentry{Spectral Residual}
\addplot[black!85, thick, dashed, mark=x, mark size=1.8pt, mark options={solid}] coordinates {(0.836,-0.2064) (1.570,-0.2015) (2.150,-0.1920) (2.714,-0.1803) (3.293,-0.1768) (3.936,-0.1674) (4.684,-0.1477) (5.478,-0.1330) (6.652,-0.0933) (8.826,-0.0042)};
\addlegendentry{Fine-Grained}
\addplot[black, dotted, thick, domain=0.28:9.03, forget plot] {0};
\draw[panelred, dashed] (axis cs:0.755,-0.215) -- (axis cs:0.755,0.008);
\node[panelred, font=\footnotesize, anchor=west, align=left, fill=white, inner sep=1pt] at (rel axis cs:0.09,0.55) {laboratory level\\0.76 degrees};
\nextgroupplot[title={\textbf{b}\,\, Laboratory gaze with\\webcam-sized error added}, xlabel={spatial error after degrading, degrees},
  ylabel={DeepGazeIIE minus center map},
  xmin=0, xmax=8.82, ymin=-0.0674, ymax=0.1372,
  legend style={font=\small, draw=none, fill=none, at={(0.5,-0.30)}, anchor=north, legend columns=1}]
\addplot[panelred, thick, mark=*, mark size=1.4pt] coordinates {(0.755,0.0772) (1.192,0.0715) (1.607,0.0636) (2.000,0.0538) (2.617,0.0380) (3.604,0.0194) (4.152,0.0079) (4.766,-0.0013) (5.771,-0.0123) (6.660,-0.0197) (8.325,-0.0204)};
\addlegendentry{constant per-viewer offset}
\addplot[panelpurple, thick, mark=triangle*, mark size=1.4pt] coordinates {(2.587,0.0772) (2.654,0.0740) (2.772,0.0685) (2.951,0.0609) (3.295,0.0478) (3.814,0.0309) (4.280,0.0199) (4.820,0.0108) (5.684,-0.0006) (6.587,-0.0077) (8.051,-0.0126)};
\addlegendentry{offset redrawn on each viewing}
\addplot[panelblue, thick, mark=square*, mark size=1.4pt] coordinates {(1.950,-0.0119) (2.193,-0.0062) (2.313,0.0006) (2.422,0.0114) (2.482,0.0204) (2.530,0.0313) (2.562,0.0482) (2.577,0.0610) (2.582,0.0684) (2.585,0.0739)};
\addlegendentry{isotropic jitter of single samples}
\addplot[black!70, thick, dashed, mark=diamond, mark size=3.0pt, mark options={solid, line width=0.8pt}] coordinates {(1.961,-0.0139) (2.200,-0.0074) (2.319,0.0001) (2.419,0.0118) (2.478,0.0212) (2.529,0.0324) (2.562,0.0493) (2.577,0.0618) (2.583,0.0689) (2.585,0.0741)};
\addlegendentry{jitter, vertical error 1.31 times horizontal}
\addplot[black, dotted, thick, domain=0:8.82, forget plot] {0};
\addplot[black!60, dashed, thick, domain=0:8.82, forget plot] {-0.0224};
\fill[panelred!14] (axis cs:3.721,-0.0674) rectangle (axis cs:3.922,0.1372);
\draw[panelred, dash dot] (axis cs:3.721,-0.0674) -- (axis cs:3.721,0.1372);
\draw[panelred, dash dot] (axis cs:3.922,-0.0674) -- (axis cs:3.922,0.1372);
\pgfplotsextra{\node[font=\footnotesize, anchor=north west, black!60, fill=white, inner sep=1pt] at (rel axis cs:0.03,0.2199) {webcam corpus $-0.022$};}
\addplot[black!45, dashed, thick, domain=0:8.82, forget plot] {0.0772};
\pgfplotsextra{\node[font=\footnotesize, anchor=north west, fill=white, inner sep=1.5pt] at (rel axis cs:0.03,0.99) {laboratory gaze, $+0.077$};}
\end{groupplot}
\end{tikzpicture}
\par\smallskip
{\scriptsize\justifying\textit{Notes.} \textbf{a} asks whether the center map wins because the webcam is imprecise. Viewings are sorted into 10 equal groups by the stable error of the viewer who produced them, and each point is the raw score of a model minus that of the center map, averaged within a group, over all devices. A value below zero means the center map wins. Median errors run from~0.84~to~8.83~degrees, and the dashed line marks the laboratory level of~0.76~degrees. In the least precise tenth the center map itself falls to chance, 0.499, so the lag of every model shrinks there. \textbf{b} asks how much spatial error it takes to turn the laboratory result into the webcam result. Laboratory gaze from MIT1003 is degraded by four forms of spatial error, and the horizontal axis gives the size of that error under the calculation that produced the webcam values. The dot-dashed lines stand at the webcam values, 3.72~and 3.92~degrees, and the dashed lines at the laboratory value of~$+0.077$ and the webcam value of~$-0.022$.\par}
\end{figure}

\begin{table}[H]
\centering
\caption{A thin content margin over the center.}
\label{tab:center_increment}
\small
{\footnotesize\textbf{a. Agreement with recorded gaze in the center-corrected score}}
\par\smallskip
\begin{tabular}{llrrrr}
\toprule
 & & \multicolumn{2}{c}{Raw gaze samples} & \multicolumn{2}{c}{Extracted fixations} \\
\cmidrule(lr){3-4}\cmidrule(l){5-6}
Model & Family & Score & Over center & Score & Over center \\
\midrule
DeepGazeIIE       & deep      & 0.552 & $+0.049$ & 0.561 & $+0.062$ \\
TranSalNet-Dense  & deep      & 0.560 & $+0.057$ & 0.571 & $+0.072$ \\
TranSalNet-Res    & deep      & 0.558 & $+0.055$ & 0.569 & $+0.070$ \\
UNISAL            & deep      & 0.558 & $+0.055$ & 0.570 & $+0.071$ \\
Spectral Residual & classical & 0.524 & $+0.021$ & 0.529 & $+0.030$ \\
Fine-Grained      & classical & 0.515 & $+0.012$ & 0.517 & $+0.017$ \\
\midrule
Center map        & baseline  & 0.503 &          & 0.499 &          \\
\bottomrule
\end{tabular}
\par\medskip
{\footnotesize\textbf{b. Reference panel of real viewers by panel size, fixed stimulus pool}}
\par\smallskip
\sbox{\tabbox}{\begin{tabular}{@{}rr@{\hspace{1.5em}}l rrr rrr@{}}
\toprule
\multicolumn{2}{c}{Panel of real viewers} & & \multicolumn{3}{c}{Extracted fixations} & \multicolumn{3}{c}{Raw gaze samples} \\
\cmidrule(r){1-2}\cmidrule(lr){4-6}\cmidrule(l){7-9}
\shortstack[r]{Pool\\viewers $n$} & \shortstack[r]{Panel\\score} & Predictor & Score & \shortstack[r]{Over\\center} & Share & Score & \shortstack[r]{Over\\center} & Share \\
\midrule
1   & 0.517 & Center map           & 0.502 &          &        & 0.499 &          &        \\
8   & 0.533 & DeepGazeIIE          & 0.536 & $+0.034$ & 71\%  & 0.527 & $+0.028$ & 78\%  \\
32  & 0.542 & TranSalNet-Dense     & 0.541 & $+0.040$ & 81\%  & 0.533 & $+0.034$ & 94\%  \\
128 & 0.549 & TranSalNet-Res       & 0.541 & $+0.039$ & 80\%  & 0.532 & $+0.033$ & 92\%  \\
256 & 0.550 & UNISAL               & 0.553 & $+0.051$ & 106\% & 0.539 & $+0.040$ & 110\% \\
    &       & Spectral Residual    & 0.525 & $+0.024$ & 49\%  & 0.520 & $+0.021$ & 57\%  \\
    &       & Fine-Grained         & 0.515 & $+0.014$ & 28\%  & 0.513 & $+0.014$ & 40\%  \\
    &       & Panel of~256 viewers & 0.550 & $+0.049$ & 100\% & 0.535 & $+0.036$ & 100\% \\
\bottomrule
\end{tabular}}%
\ifdim\wd\tabbox>\textwidth\resizebox{\textwidth}{!}{\usebox{\tabbox}}\else\usebox{\tabbox}\fi
\par\smallskip
{\scriptsize\justifying\settabnotew\noindent\parbox{\tabnotew}{\setlength{\emergencystretch}{2em}\textit{Notes.} \textbf{a}, the center-corrected score of each model against recorded gaze, on raw samples and on extracted fixations, one score per image, with each map divided by the model's average map (Methods). The margin over center is the model's score minus the center map's, taken on unrounded values, so a margin can differ by~0.001 from the difference of the rounded entries, and the center map's own row leaves both margin cells empty. \textbf{b}, the reference panel of Methods. A map built from~$n$ viewers of an image is scored on viewers held out of it, on extracted fixations, and each model's released map is scored on the same held-out viewers of the same 35 images on both signals. The share is the model's margin over center divided by the margin of the panel of 256, (model $-$ center) / (panel of 256 $-$ center), taken on unrounded values, so 100\% means the model matches a panel of 256 people.\par}}
\end{table}

\begin{table}[H]
\centering
\caption{Each gap re-estimated under every specification and check.}
\label{tab:spec_grid}
{\footnotesize\textbf{a. Specification grid, all devices}}\par\smallskip
\begingroup
\footnotesize
\setlength{\tabcolsep}{4pt}
\sbox{\tabbox}{\begin{tabular}{@{}l rrrrrrr@{}}
\toprule
Gap & S0 naive & S1 intercepts & S2 controls & S3 weights & S4 exclusion & S5 distance & S6 both \\
\midrule
Age (young $-$ old)        & $+0.013^{***}$ & $+0.014^{***}$ & $+0.014^{***}$ & $+0.015^{***}$ & $+0.020^{***}$ & $+0.015^{***}$ & $+0.015^{***}$ \\
Race (Black $-$ White)     & $-0.005^{***}$ & $-0.005$~($p=0.06$) & $-0.006^{*}$ & $-0.006^{*}$ & $-0.008^{*}$ & $-0.006^{*}$ & $-0.007^{*}$ \\
Race (Asian $-$ White)     & $+0.000$       & $-0.002$       & $-0.002$       & $-0.002$       & $-0.004$ & $-0.002$ & $-0.002$ \\
Gender (women $-$ men)     & $+0.003^{**}$  & $+0.004^{*}$   & $+0.004^{*}$   & $+0.004^{*}$   & $+0.002$~($p=0.30$) & $+0.003$~($p=0.05$) & $+0.004^{*}$ \\
Extremity (extreme $-$ moderate) & $-0.004^{***}$ & $-0.006^{***}$ & $-0.006^{**}$  & $-0.006^{***}$ & $-0.006^{**}$ & $-0.006^{***}$ & $-0.006^{***}$ \\
Income (top $-$ bottom)    & $+0.003^{**}$  & $+0.000$       & $+0.001$       & $+0.001$       & $+0.002$ & $+0.000$ & $+0.001$ \\
Education (top $-$ bottom) & $+0.002$       & $-0.001$       & $-0.001$       & $-0.001$       & $-0.001$ & $-0.001$ & $-0.001$ \\
Party (Dem $-$ Rep, unified) & \multicolumn{7}{c}{$-0.001$~to~$-0.000$, $p$~from~0.28~to~0.85 in every specification} \\
Ideology (lib $-$ cons)    & $+0.000$       & $-0.001$       & $-0.000$       & $-0.001$       & $-0.000$ & $-0.001$ & $-0.000$ \\
\bottomrule
\end{tabular}}%
\resizebox{\textwidth}{!}{\usebox{\tabbox}}
\par\smallskip
{\scriptsize\justifying\settabnotew\noindent\parbox{\tabnotew}{\setlength{\emergencystretch}{2em}\textit{Notes.} Each row is a gap between two groups in UNISAL's accuracy on raw gaze samples, and each column estimates it a different way, so a gap that keeps its sign and size across the columns holds under every choice of analysis. Cells give the gap with its $p$ value. S0 pools all viewings without intercepts, with unclustered standard errors. S1 adds intercepts for each combination of recruitment wave and device and for each image, so that groups are compared on the same images under the same conditions, and clusters standard errors by participant. S2 is S1 with controls for recording quality. S3 is S1 with each viewing weighted by its recording quality. S4 is S1 on the subsample that leaves out smartphones and keeps viewings with at least 90\%~of their gaze on screen. S5 is S1 with each viewer's distance from the crowd, the stable spatial error of Methods, standardized, entered as a regressor and in interaction with the group indicator, so that groups are compared at equal recording error, and S6 is S2 with the same two terms. Stars mark $^{*}p<0.05$, $^{**}p<0.01$~and $^{***}p<0.001$. A cell printed as $\pm0.000$ has a magnitude below 0.0005, with its sign shown.\par}}
\endgroup
\par\medskip
{\footnotesize\textbf{b. Desktop battery on the released maps}}\par\smallskip
\begingroup
\small
\sbox{\tabbox}{\begin{tabular}{@{}l rrr@{}}
\toprule
Check & Age (young $-$ old) & Race (Black $-$ White) & Gender (women $-$ men) \\
\midrule
Base, intercepts and clustering    & $+0.018$~($p<10^{-5}$) & $-0.007$~($p=0.057$) & $+0.002$~($p=0.47$) \\
Fixation signal                    & $+0.020$~($p<10^{-5}$) & $-0.009$~($p=0.049$) & $+0.002$~($p=0.59$) \\
Full trait controls                & $+0.018$~($p<10^{-5}$) & $-0.009$~($p=0.019$) & $+0.001$~($p=0.72$) \\
Quality controls                   & $+0.018$~($p<10^{-5}$) & $-0.008$~($p=0.049$) & $+0.003$~($p=0.29$) \\
Quality weights                    & $+0.019$~($p<10^{-4}$) & $-0.008$~($p=0.051$) & $+0.002$~($p=0.48$) \\
Gaze-on-screen exclusion           & $+0.022$~($p<10^{-5}$) & $-0.009$~($p=0.030$) & $+0.001$~($p=0.72$) \\
Clustering by person and image     & $+0.018$~($p<10^{-4}$) & $-0.007$~($p=0.050$) & $+0.002$~($p=0.46$) \\
Uncorrected score as outcome       & $+0.023$~($p=0.004$)   & $-0.019$~($p=0.032$) & $-0.006$~($p=0.28$) \\
Wave 1 only                        & $+0.022$~($p=0.001$)   & $-0.015$~($p=0.028$) & $+0.013$~($p=0.010$) \\
Wave 2 only                        & $+0.018$~($p=0.0002$)  & $-0.013$~($p=0.019$) & $-0.003$~($p=0.45$) \\
Wave 3 only                        & $+0.012$~($p=0.12$)    & $+0.007$~($p=0.30$) & $+0.006$~($p=0.21$) \\
Wave 4 only, four topics           & $+0.020$~($p=0.001$)   & $-0.008$~($p=0.33$) & $+0.002$~($p=0.61$) \\
Rank test across images            & $p=2\times10^{-6}$     & $p=0.0012$ & $p=0.21$ \\
Equal-size group subsample         & CI $[+0.016, +0.021]$  & CI $[-0.016, -0.002]$ & CI $[+0.001, +0.003]$ \\
Each image dropped in turn, range  & $[+0.017, +0.019]$     & $[-0.008, -0.006]$ & $[+0.0015, +0.0024]$ \\
Redrawn comparison points          & unchanged              & unchanged & unchanged \\
\bottomrule
\end{tabular}}%
\resizebox{\textwidth}{!}{\usebox{\tabbox}}
\par\smallskip
{\scriptsize\justifying\settabnotew\noindent\parbox{\tabnotew}{\setlength{\emergencystretch}{2em}\textit{Notes.} The battery repeats the three gaps among desktop viewers of waves 1 to 3 under one change at a time, and the wave 4 row uses every viewer of wave 4 with intercepts for the device and the image. Gaze enters as raw samples unless a row states otherwise. The outcome is UNISAL's margin over the center map on the same gaze, and the row marked uncorrected score uses the model score itself. Every row is a separate re-estimate. A negative race value means the model predicts Black viewers less accurately. CI is a 95\% confidence interval.\par}}
\endgroup
\end{table}

\begin{table}[H]
\centering
\caption{Model accuracy by group of viewers in the public datasets.}
\label{tab:public_models}
\footnotesize
\sbox{\tabbox}{\begin{tabular}{@{}ll r rrr@{}}
\toprule
Dataset & Comparison, first group minus second & Viewers & UNISAL & Center map & UNISAL minus center \\
\midrule
Museum visitors & Aged 19 to 34 minus 55 and older & 1{,}820 and 909 & $+0.45$ (0.29) & $-3.46$ (0.30) & $+3.91$ (0.30) \\
 & Aged 19 to 34 minus 35 to 54 & 1{,}820 and 2{,}084 & $+0.82$ (0.21) & $+0.09$ (0.22) & $+0.73$ (0.21) \\
 & Aged 19 to 34 minus 9 and younger & 1{,}820 and 2{,}164 & $-1.61$ (0.24) & $-7.93$ (0.24) & $+6.32$ (0.24) \\
 & Aged 19 to 34 minus 10 to 18 & 1{,}820 and 4{,}856 & $+0.20$ (0.20) & $-3.34$ (0.20) & $+3.54$ (0.21) \\
 & Women minus men & 5{,}113 and 6{,}345 & $-1.30$ (0.14) & $-0.21$ (0.16) & $-1.09$ (0.15) \\
 & Without glasses minus with glasses & 8{,}565 and 3{,}442 & $+1.29$ (0.16) & $+1.61$ (0.16) & $-0.32$ (0.16) \\
\addlinespace
Six countries & Women minus men & 216 and 190 & $-1.39$ (0.50) & $-2.79$ (0.66) & $+1.40$ (0.43) \\
 & University minus secondary education & 250 and 126 & $-0.72$ (0.50) & $-0.97$ (0.69) & $+0.26$ (0.46) \\
 & Upper middle minus middle class & 77 and 275 & $-2.64$ (0.82) & $-2.35$ (0.92) & $-0.29$ (0.53) \\
 & Middle minus lower class & 275 and 46 & $+0.33$ (0.70) & $+1.60$ (0.89) & $-1.27$ (0.55) \\
 & Czechia minus other samples & 92 and 316 & $+0.50$ (0.51) & $+0.57$ (0.73) & $-0.07$ (0.49) \\
 & Taiwan minus other samples & 65 and 343 & $-6.01$ (0.84) & $-5.73$ (0.94) & $-0.28$ (0.53) \\
 & Turkey, east, minus other samples & 57 and 351 & $+2.64$ (0.70) & $+4.40$ (1.12) & $-1.76$ (0.85) \\
 & Bulgaria minus other samples & 56 and 352 & $+0.40$ (0.52) & $-1.05$ (0.77) & $+1.45$ (0.57) \\
 & Germany minus other samples & 54 and 354 & $-0.36$ (0.53) & $-1.35$ (0.79) & $+0.98$ (0.56) \\
 & Ghana minus other samples & 42 and 366 & $+2.37$ (0.67) & $+2.74$ (0.94) & $-0.37$ (0.59) \\
 & Turkey, west, minus other samples & 42 and 366 & $+1.92$ (0.69) & $+1.81$ (0.91) & $+0.11$ (0.62) \\
\addlinespace
Wilming and colleagues & Children minus older adults & 18 and 17 & $+3.11$ (2.15) & $-0.44$ (3.51) & $+3.55$ (1.65) \\
 & Children minus young adults & 18 and 23 & $+9.27$ (1.59) & $+9.72$ (2.17) & $-0.45$ (0.87) \\
 & Young adults minus older adults & 23 and 17 & $-6.18$ (2.24) & $-10.21$ (3.66) & $+4.02$ (1.68) \\
\addlinespace
Museum, one image & Aged 18 to 34 minus 55 and older & 2{,}142 and 186 & $-0.28$ (0.77) & $+0.39$ (0.67) & $-0.67$ (0.72) \\
 & Women minus men & 1{,}792 and 2{,}325 & $-0.82$ (0.33) & $+0.27$ (0.27) & $-1.09$ (0.30) \\
\addlinespace
News pages & Median age or younger minus older & 14 and 11 & $-1.49$ (0.85) & $-1.11$ (1.09) & $-0.38$ (0.97) \\
 & Men minus women & 12 and 13 & $+0.76$ (0.89) & $+0.86$ (1.07) & $-0.10$ (1.01) \\
\addlinespace
Young viewers, videos & Aged 15 and older minus 5 to 10 & 284 and 1{,}215 & & & $-0.34$ (0.31) \\
 & Aged 11 to 14 minus 5 to 10 & 816 and 1{,}215 & & & $+0.02$ (0.20) \\
 & Girls minus boys & 828 and 1{,}487 & & & $+0.26$ (0.19) \\
\bottomrule
\end{tabular}}%
\resizebox{\textwidth}{!}{\usebox{\tabbox}}
\par\smallskip
{\scriptsize\justifying\settabnotew\noindent\parbox{\tabnotew}{\setlength{\emergencystretch}{2em}\textit{Notes.} Each entry is a difference between two groups in the center-corrected score, in correct rankings per 100 pairs, with standard errors clustered by viewer in parentheses. A negative entry means the map predicts the first group less accurately. The last column is the difference in UNISAL's margin over the center map, which removes the part of the gap that the center map shares. Museum visitors are the dataset of Linka and colleagues \autocite{linka2025protracted}, 12{,}007 scored viewers of 40~images, where UNISAL scores 80.3 and the center map 50.7. Six countries are the students of \v{C}en\v{e}k and colleagues \autocite{cenek2025cross}, 408 viewers of 70~images, with scores of 73.6 and 50.6. The data of Wilming and colleagues \autocite{wilming2017extensive} cover 58 viewers of the 249 of its 255~images whose files are in the released archive, with scores of 61.7 and 51.6, and its children and young adults have mean ages of 7.6 and 22.1 years and its older adults 80.6. The museum study with one image is from Strauch and colleagues \autocite{strauch2023saliency}, 6{,}909 viewers on the consumer eye tracker, with scores of 60.6 and 51.4. Its two group rows use the 4,117 visitors with valid demographics, leaving out visitors who kept the default year of birth 2000 or the default gender other (Extended Data Table~\ref{tab:sample_composition}b). News pages are the FakeNewsPerception dataset \autocite{sumer2021fakenewsperception}, 25 viewers of 119~pages, the true and false versions of 60 news items less one page missing from the release, with scores of 56.2 and 50.3. Young viewers are the children and adolescents of the Healthy Brain Network \autocite{alexander2017open}. For their videos the table gives the mean margin of the four deep networks over the center map, which averages 14.3 points.\par}}
\end{table}

\begin{table}[H]
\centering
\caption{A network trained on a group, and the gain of a group's own pool, in every dataset.}
\label{tab:group_readouts}
\footnotesize
\setlength{\tabcolsep}{3.5pt}
\sbox{\tabbox}{\begin{tabular}{@{}ll r r rr r r r@{}}
\toprule
 & & \multicolumn{2}{c}{UNISAL} & \multicolumn{3}{c}{Readout trained on} & \multicolumn{2}{c}{Pool of real viewers} \\
\cmidrule(lr){3-4}\cmidrule(lr){5-7}\cmidrule(lr){8-9}
Axis & Group & Score & First minus second & Own group & Mixed & Own minus mixed & Own group & Own minus mixed \\
\midrule
\multicolumn{9}{@{}l}{\textit{National sample, news images}} \\
Age & 18 to 34 & 55.0 & $+3.68$ (1.01) & 54.8 & 54.1 & $+0.71$ (0.17) & 54.2 & $+0.87$ (0.43) \\
 & 55 and older & 51.3 &  & 50.4 & 50.6 & $-0.22$ (0.25) & 51.3 & $-0.13$ (0.32) \\
Race & Black & 51.0 & $-2.16$ (1.24) & 50.2 & 51.0 & $-0.81$ (0.35) & 50.2 & $-0.20$ (0.61) \\
 & White & 53.2 &  & 52.6 & 52.6 & $-0.02$ (0.15) & 52.4 & $-0.06$ (0.26) \\
Gender & women & 52.4 & $-1.12$ (0.73) & 52.1 & 51.9 & $+0.14$ (0.06) & 52.2 & $+0.56$ (0.25) \\
 & men & 53.5 &  & 52.8 & 52.8 & $-0.00$ (0.09) & 53.1 & $-0.05$ (0.25) \\
Conviction & strong conviction & 52.4 & $-1.43$ (0.73) & 51.8 & 52.2 & $-0.40$ (0.12) & 52.1 & $+0.04$ (0.16) \\
 & moderate & 53.9 &  & 53.8 & 53.3 & $+0.48$ (0.22) & 53.1 & $+0.30$ (0.22) \\
Position on the issues & leans Republican & 52.0 & $-1.73$ (1.67) & 51.5 & 51.9 & $-0.37$ (0.19) & 51.6 & $+0.22$ (0.65) \\
 & leans Democratic & 53.7 &  & 53.4 & 53.4 & $-0.04$ (0.12) & 53.7 & $+0.14$ (0.46) \\
\addlinespace
\multicolumn{9}{@{}l}{\textit{Museum visitors, scenes}} \\
Age, adults & 19 to 34 & 80.2 & $+0.47$ (0.40) & 82.1 & 82.1 & $-0.05$ (0.08) & 84.4 & $+0.65$ (0.07) \\
 & 55 and older & 79.8 &  & 80.3 & 80.4 & $-0.12$ (0.20) & 83.0 & $-0.31$ (0.07) \\
Age, children & 19 to 34 & 80.3 & $-1.54$ (0.63) & 82.2 & 82.2 & $-0.03$ (0.06) & 84.6 & $+1.76$ (0.14) \\
 & 9 and younger & 81.9 &  & 81.7 & 81.3 & $+0.35$ (0.18) & 83.5 & $-0.89$ (0.09) \\
Gender, adults & women & 79.4 & $-1.27$ (0.26) & 80.7 & 80.9 & $-0.12$ (0.08) & 83.6 & $+0.12$ (0.04) \\
 & men & 80.6 &  & 82.2 & 82.2 & $+0.03$ (0.06) & 85.0 & $+0.00$ (0.03) \\
\addlinespace
\multicolumn{9}{@{}l}{\textit{Students of six countries, scenes}} \\
Gender & women & 73.2 & $-1.40$ (0.55) & 73.9 & 74.0 & $-0.13$ (0.12) & 75.7 & $+0.73$ (0.09) \\
 & men & 74.6 &  & 74.8 & 74.7 & $+0.11$ (0.11) & 76.4 & $-0.52$ (0.07) \\
Education & without university & 74.1 & $+0.72$ (0.53) & 74.8 & 74.8 & $-0.01$ (0.14) & 76.7 & $+0.19$ (0.08) \\
 & university & 73.3 &  & 74.1 & 73.8 & $+0.26$ (0.10) & 75.3 & $-0.05$ (0.08) \\
\addlinespace
\multicolumn{9}{@{}l}{\textit{Wilming and colleagues, scenes}} \\
Children against older adults & children & 66.7 & $+3.39$ (2.15) & 65.3 & 65.1 & $+0.13$ (0.21) & 69.3 & $+1.77$ (0.44) \\
 & older adults & 63.3 &  & 60.1 & 60.3 & $-0.20$ (0.17) & 63.5 & $-0.37$ (0.63) \\
Young against older adults & young adults & 57.1 & $-6.20$ (2.17) & 60.0 & 59.7 & $+0.36$ (0.17) & 62.2 & $+2.32$ (0.43) \\
 & older adults & 63.3 &  & 60.4 & 60.2 & $+0.23$ (0.25) & 63.6 & $-0.31$ (0.68) \\
Children against young adults & children & 66.7 & $+9.59$ (1.59) & 65.3 & 65.1 & $+0.18$ (0.21) & 69.0 & $-0.07$ (0.43) \\
 & young adults & 57.1 &  & 60.1 & 60.0 & $+0.05$ (0.13) & 62.6 & $+1.50$ (0.40) \\
\bottomrule
\end{tabular}}%
\resizebox{\textwidth}{!}{\usebox{\tabbox}}
\par\smallskip
{\scriptsize\justifying\settabnotew\noindent\parbox{\tabnotew}{\setlength{\emergencystretch}{2em}\textit{Notes.} The table asks whether a map made for a group predicts that group better than a map made for everyone. Entries are correct rankings per 100 pairs in the center-corrected score, on images held out of every fit and averaged over images, with standard errors in parentheses from 1,000 bootstrap draws that resample the images and, within each group, the viewers with all of their images. The UNISAL columns score the released map on the group's viewers, and the first-minus-second column gives its accuracy on the first group of the axis minus the second. The readout columns compare a network trained on the group's own pool with the same network trained on a mixed pool of the same size, in five folds by image, and the own-minus-mixed column gives the gain from training on the group. The readout keeps VGG-16 \autocite{simonyan2015vgg} as pretrained, takes its relu4\_3 and relu5\_3 layers, and trains a head of three stages that reduces 1024 channels to 64 and then to 16, with a learned center and the logarithms of the UNISAL and DeepGazeIIE maps as starting points, by a pairwise ranking loss with early stopping on an inner fifth of the training images. The pool of real viewers is a map built from half of the group's viewers, or from an equally large mixed pool, and scored on the other half. The split of a group into the pool and the viewers it is scored on stays fixed across the bootstrap draws. Strong conviction is a viewer at points 1, 2, 6, 7 of the 7-point ideology scale, and moderate a viewer at 3 to 5. A group's row uses the held-out images on which both groups of the pair supply five or more viewers, and the pool of real viewers is half of the smaller group on the image, 19 to 173 viewers in the national sample. The rows on the position on the issues use the 48 images of wave 4. Country is a contrast of trackers, so its accuracy appears in Extended Data Table~\ref{tab:public_models} alone.\par}}
\end{table}

\begin{table}[H]
\centering
\caption{A model of groups. The group of a viewer predicted from how long the viewer looks, and the group of an audience from a few viewers.}
\label{tab:group_model}
\scriptsize
\setlength{\tabcolsep}{3.5pt}
\renewcommand{\arraystretch}{0.92}
\textbf{a.} One held-out viewer\par\smallskip
\sbox{\tabbox}{\begin{tabular}{@{}ll r rrr rr@{}}
\toprule
 & & & \multicolumn{3}{c}{Model trained on how long viewers look} & \multicolumn{2}{c}{Model trained on where} \\
\cmidrule(lr){4-6}\cmidrule(lr){7-8}
First group & Second group & Viewers & Balanced accuracy & AUC & Shuffled labels & Balanced accuracy & AUC \\
\midrule
\multicolumn{8}{@{}l}{\textit{Museum visitors, scenes}} \\
9 and younger & 19 to 34 & 4,000 & 0.87 (0.01) & 0.93 (0.00) & 0.50 & 0.78 (0.01) & 0.84 (0.01) \\
12 and younger & 25 to 54 & 8,001 & 0.86 (0.00) & 0.93 (0.00) & 0.50 & 0.75 (0.01) & 0.81 (0.00) \\
19 to 34 & 55 and older & 2,735 & 0.76 (0.01) & 0.85 (0.01) & 0.50 & 0.60 (0.01) & 0.65 (0.01) \\
19 to 34 & 75 and older & 1,951 & 0.60 (0.02) & 0.83 (0.01) & 0.50 & 0.54 (0.01) & 0.70 (0.02) \\
women, adults & men, adults & 4,655 & 0.73 (0.01) & 0.81 (0.01) & 0.49 & 0.54 (0.01) & 0.56 (0.01) \\
English, adults & German, adults & 4,822 & 0.52 (0.01) & 0.59 (0.02) & 0.50 & 0.52 (0.01) & 0.55 (0.02) \\
\addlinespace
\multicolumn{8}{@{}l}{\textit{Children and adolescents, videos}} \\
5 to 8 & 15 and older & 1,232 & 0.75 (0.02) & 0.85 (0.01) & 0.49 & 0.51 (0.02) & 0.51 (0.02) \\
5 to 8 & 9 to 11 & 1,635 & 0.59 (0.01) & 0.63 (0.01) & 0.50 & 0.50 (0.01) & 0.51 (0.01) \\
5 to 10 & 11 to 14 & 2,030 & 0.61 (0.01) & 0.68 (0.01) & 0.50 & 0.51 (0.01) & 0.51 (0.01) \\
girls & boys & 2,314 & 0.55 (0.01) & 0.58 (0.01) & 0.51 & 0.50 (0.01) & 0.50 (0.01) \\
\addlinespace
\multicolumn{8}{@{}l}{\textit{Students of six countries, scenes}} \\
Ghana and Turkey & Germany and Czechia & 287 & 0.85 (0.02) & 0.92 (0.02) & 0.47 & 0.65 (0.03) & 0.72 (0.03) \\
without university & university & 408 & 0.64 (0.02) & 0.71 (0.03) & 0.46 & 0.50 (0.02) & 0.51 (0.03) \\
women & men & 406 & 0.65 (0.02) & 0.70 (0.03) & 0.52 & 0.54 (0.02) & 0.54 (0.03) \\
\addlinespace
\multicolumn{8}{@{}l}{\textit{National sample, news images}} \\
18 to 34 & 55 and older & 1,653 & 0.63 (0.01) & 0.68 (0.01) & 0.49 & 0.57 (0.01) & 0.60 (0.01) \\
Black & White & 2,641 & 0.51 (0.00) & 0.58 (0.02) & 0.50 & 0.51 (0.01) & 0.53 (0.02) \\
income brackets 1 to 4 & brackets 9 to 12 & 1,985 & 0.60 (0.01) & 0.63 (0.01) & 0.49 & 0.56 (0.01) & 0.59 (0.01) \\
education levels 1 to 4 & levels 5 to 7 & 3,019 & 0.56 (0.01) & 0.59 (0.01) & 0.49 & 0.52 (0.01) & 0.53 (0.01) \\
Democrats & Republicans & 2,546 & 0.49 (0.01) & 0.49 (0.01) & 0.50 & 0.49 (0.01) & 0.49 (0.01) \\
ideology 1 to 3, conservative & 5 to 7, liberal & 2,205 & 0.50 (0.01) & 0.49 (0.01) & 0.49 & 0.50 (0.01) & 0.50 (0.01) \\
ideology 1, 2, 6, 7, strong conviction & 3 to 5, moderate & 3,020 & 0.53 (0.01) & 0.55 (0.01) & 0.50 & 0.51 (0.01) & 0.51 (0.01) \\
leans Republican, four topics & leans Democratic & 243 & 0.56 (0.03) & 0.59 (0.04) & 0.49 & 0.57 (0.03) & 0.59 (0.04) \\
leans Republican, immigration & leans Democratic & 2,102 & 0.50 (0.01) & 0.50 (0.01) & 0.49 & 0.51 (0.01) & 0.52 (0.01) \\
thermometer gap 50 and more & 20 and less & 2,206 & 0.54 (0.01) & 0.58 (0.01) & 0.51 & 0.53 (0.01) & 0.54 (0.01) \\
\bottomrule
\end{tabular}}%
\ifdim\wd\tabbox>\textwidth\resizebox{\textwidth}{!}{\usebox{\tabbox}}\else\usebox{\tabbox}\fi
\par\medskip
\textbf{b.} Audience identification, AUC by the number of viewers drawn from each group\par\smallskip
\sbox{\tabbox}{\begin{tabular}{@{}ll rrrrrr@{}}
\toprule
First group & Second group & 1 & 3 & 5 & 10 & 20 & 50 \\
\midrule
\multicolumn{8}{@{}l}{\textit{Museum visitors, scenes}} \\
9 and younger & 19 to 34 & 0.94 [0.93, 0.95] & 0.99 [0.99, 1.00] & 1.00 [1.00, 1.00] & 1.00 [1.00, 1.00] & 1.00 [1.00, 1.00] & 1.00 [1.00, 1.00] \\
19 to 34 & 55 and older & 0.86 [0.84, 0.88] & 0.96 [0.95, 0.97] & 0.99 [0.98, 0.99] & 1.00 [1.00, 1.00] & 1.00 [1.00, 1.00] & 1.00 [1.00, 1.00] \\
\addlinespace
\multicolumn{8}{@{}l}{\textit{Children and adolescents, videos}} \\
5 to 8 & 15 and older & 0.87 [0.83, 0.90] & 0.97 [0.95, 0.98] & 0.99 [0.98, 1.00] & 1.00 [1.00, 1.00] & 1.00 [1.00, 1.00] & 1.00 [1.00, 1.00] \\
\addlinespace
\multicolumn{8}{@{}l}{\textit{National sample, news images}} \\
18 to 34 & 55 and older & 0.65 [0.62, 0.67] & 0.74 [0.69, 0.77] & 0.78 [0.74, 0.82] & 0.86 [0.79, 0.91] & 0.93 [0.87, 0.97] &  \\
Black & White & 0.56 [0.53, 0.59] & 0.60 [0.56, 0.65] & 0.63 [0.56, 0.69] & 0.67 [0.57, 0.74] & 0.70 [0.60, 0.81] &  \\
income brackets 1 to 4 & brackets 9 to 12 & 0.54 [0.52, 0.57] & 0.57 [0.52, 0.62] & 0.59 [0.53, 0.64] & 0.63 [0.56, 0.70] & 0.67 [0.57, 0.76] &  \\
education levels 1 to 4 & levels 5 to 7 & 0.55 [0.53, 0.57] & 0.58 [0.54, 0.61] & 0.60 [0.56, 0.64] & 0.64 [0.58, 0.69] & 0.68 [0.61, 0.75] & 0.73 [0.63, 0.81] \\
Democrats & Republicans & 0.49 [0.47, 0.51] & 0.48 [0.44, 0.52] & 0.47 [0.43, 0.53] & 0.47 [0.39, 0.54] & 0.45 [0.38, 0.56] & 0.43 [0.31, 0.57] \\
ideology 1 to 3, conservative & 5 to 7, liberal & 0.48 [0.46, 0.51] & 0.47 [0.44, 0.51] & 0.47 [0.41, 0.51] & 0.45 [0.39, 0.52] & 0.43 [0.34, 0.54] & 0.40 [0.28, 0.53] \\
ideology 1, 2, 6, 7, strong conviction & 3 to 5, moderate & 0.53 [0.51, 0.55] & 0.54 [0.52, 0.59] & 0.56 [0.52, 0.61] & 0.58 [0.52, 0.65] & 0.62 [0.52, 0.68] & 0.67 [0.55, 0.78] \\
leans Republican, four topics & leans Democratic & 0.60 [0.54, 0.67] & 0.65 [0.52, 0.76] & 0.70 [0.54, 0.82] & 0.76 [0.55, 0.92] & 0.83 [0.54, 0.96] &  \\
leans Republican, immigration & leans Democratic & 0.50 [0.48, 0.52] & 0.49 [0.45, 0.54] & 0.50 [0.44, 0.56] & 0.50 [0.42, 0.58] & 0.50 [0.40, 0.60] & 0.51 [0.36, 0.65] \\
thermometer gap 50 and more & 20 and less & 0.54 [0.51, 0.56] & 0.56 [0.51, 0.60] & 0.57 [0.52, 0.64] & 0.60 [0.53, 0.68] & 0.65 [0.55, 0.73] &  \\
\bottomrule
\end{tabular}}%
\resizebox{\textwidth}{!}{\usebox{\tabbox}}
\par\smallskip
{\scriptsize\justifying\noindent\parbox{\textwidth}{\setlength{\emergencystretch}{2em}\textit{Notes.} \textbf{a}, a gradient-boosting classifier predicts the group of viewers held out of its training, in five folds repeated three times, with every reference built from the training viewers and, in the national sample, within the recruitment wave and device class. Balanced accuracy averages the share predicted correctly in each group, and AUC is the chance that a viewer of the first group ranks above a viewer of the second, both 0.5 at chance, with standard errors in parentheses. Shuffled labels repeats the procedure after the group labels are shuffled. The place model sees only where the fixations land. \textbf{b}, each viewer's held-out probability is averaged over $n$ viewers drawn from each group, in the national sample from the same recruitment wave, and the AUC compares the two averages over 4,000 draws, with 95\% intervals from 200 bootstrap resamples. With shuffled labels the cells lie between 0.47 and 0.55, except in the national sample, where they reach 0.62 for age and 0.61 for race at 10 viewers. A cell is empty where $n$ exceeds half the smaller group.\par}}
\end{table}

\begin{table}[H]
\centering
\caption{Every contrast in the time per fixation.}
\label{tab:rest_time_all}
\footnotesize
\setlength{\tabcolsep}{4pt}
\sbox{\tabbox}{\begin{tabular}{@{}lll rr r c r@{}}
\toprule
 & & & \multicolumn{2}{c}{Time per fixation, ms} & & & \\
\cmidrule(lr){4-5}
Dataset and content & First group & Second group & First & Second & Difference & 95\% interval & Viewers \\
\midrule
\multicolumn{8}{@{}l}{\textit{Age}} \\
Museum visitors, scenes & 75 and older & 19 to 34 & 396.9 & 323.9 & $+22.5\%$ & $+15.9$ to $+29.2$ & 126 and 1{,}823 \\
Museum visitors, scenes & 3 to 6 & 19 to 34 & 383.6 & 323.9 & $+18.5\%$ & $+14.9$ to $+22.1$ & 378 and 1{,}823 \\
Museum, calibration under 1$^{\circ}$, without glasses & 75 and older & 19 to 34 & 357.3 & 302.3 & $+18.2\%$ & $+6.1$ to $+31.1$ & 30 and 1{,}006 \\
Young viewers, videos & 5 to 7 & 14 to 17 & 277.6 & 238.9 & $+16.2\%$ & $+10.9$ to $+21.8$ & 559 and 253 \\
Wilming and colleagues, scenes & Children & Young adults & 400.6 & 294.9 & $+35.9\%$ & $+17.9$ to $+56.1$ & 18 and 23 \\
Wilming and colleagues, scenes & Older adults & Young adults & 279.0 & 294.9 & $-5.4\%$ & $-18.7$ to $+9.7$ & 17 and 23 \\
National, four topics & 55 and older & 18 to 34 & 547.2 & 437.4 & $+25.1\%$ & $+13.1$ to $+38.4$ & 132 and 76 \\
National, all news & 55 and older & 18 to 34 & 507.5 & 479.5 & $+5.8\%$ & $+2.3$ to $+9.3$ & 766 and 883 \\
\addlinespace
\multicolumn{8}{@{}l}{\textit{Race}} \\
University, videos & Black & White & 424.0 & 379.9 & $+11.6\%$ & $+0.7$ to $+22.5$ & 29 and 177 \\
University, videos & Hispanic & White & 399.5 & 379.9 & $+5.2\%$ & $-1.6$ to $+12.6$ & 80 and 177 \\
National, all news & Black & White & 542.8 & 484.8 & $+12.0\%$ & $+7.3$ to $+16.8$ & 335 and 2{,}302 \\
National, four topics & Black & White & 549.7 & 492.8 & $+11.5\%$ & $-0.3$ to $+23.6$ & 44 and 295 \\
National, immigration & Black & White & 541.5 & 483.2 & $+12.1\%$ & $+7.3$ to $+17.2$ & 291 and 2{,}007 \\
\addlinespace
\multicolumn{8}{@{}l}{\textit{Position on the issues}} \\
National, four topics & Leans Republican & Leans Democratic & 556.8 & 464.8 & $+19.8\%$ & $+9.1$ to $+30.6$ & 80 and 162 \\
National, four topics & Republican, over 1 step & Democratic, over 1 step & 569.6 & 462.4 & $+23.2\%$ & $+10.6$ to $+37.1$ & 45 and 132 \\
National, immigration & Toward Republican & Toward Democratic & 495.4 & 485.0 & $+2.2\%$ & $-0.9$ to $+5.2$ & 1{,}202 and 893 \\
\addlinespace
\multicolumn{8}{@{}l}{\textit{Country}} \\
Six countries, scenes & Ghana and Turkey & Germany and Czechia & 328.0 & 293.4 & $+11.8\%$ & $+7.9$ to $+16.2$ & 141 and 146 \\
\addlinespace
\multicolumn{8}{@{}l}{\textit{Language}} \\
Museum adults, scenes & English & German & 357.1 & 325.5 & $+9.7\%$ & $+5.5$ to $+14.2$ & 369 and 4{,}449 \\
\addlinespace
\multicolumn{8}{@{}l}{\textit{Income}} \\
National, all news & Brackets 1 to 4 & Brackets 9 to 12 & 513.5 & 477.0 & $+7.7\%$ & $+4.7$ to $+11.2$ & 911 and 1{,}069 \\
National, four topics & Brackets 1 to 4 & Brackets 9 to 12 & 494.8 & 476.2 & $+3.9\%$ & $-4.8$ to $+12.8$ & 89 and 182 \\
Six countries, scenes & Lower, middle class & Upper middle, upper & 310.3 & 307.4 & $+0.9\%$ & $-2.6$ to $+4.5$ & 324 and 84 \\
\addlinespace
\multicolumn{8}{@{}l}{\textit{Ideology}} \\
National, all news & Most conservative & Most liberal & 502.3 & 467.1 & $+7.5\%$ & $+2.2$ to $+13.3$ & 306 and 384 \\
National, four topics & Most conservative & Most liberal & 545.7 & 463.2 & $+17.8\%$ & $+3.0$ to $+33.7$ & 43 and 76 \\
National, all news & Conservative, 1 to 3 & Liberal, 5 to 7 & 502.4 & 482.2 & $+4.2\%$ & $+1.3$ to $+7.3$ & 1{,}033 and 1{,}167 \\
National, four topics & Conservative, 1 to 3 & Liberal, 5 to 7 & 515.9 & 482.3 & $+6.9\%$ & $-0.6$ to $+15.2$ & 158 and 191 \\
National, all news & Points 1 and 7 & Point 4 & 482.7 & 498.6 & $-3.2\%$ & $-6.4$ to $+0.4$ & 690 and 812 \\
National, four topics & Points 1 and 7 & Point 4 & 493.0 & 501.7 & $-1.7\%$ & $-12.0$ to $+10.2$ & 119 and 50 \\
\addlinespace
\multicolumn{8}{@{}l}{\textit{Education}} \\
National, all news & Levels 1 to 4 & Levels 5 to 7 & 499.7 & 486.8 & $+2.6\%$ & $+0.1$ to $+5.2$ & 1{,}561 and 1{,}450 \\
National, four topics & Levels 1 to 4 & Levels 5 to 7 & 531.9 & 470.7 & $+13.0\%$ & $+6.0$ to $+20.4$ & 178 and 221 \\
Six countries, scenes & Without university & University & 302.8 & 314.0 & $-3.6\%$ & $-6.6$ to $-0.4$ & 158 and 250 \\
\addlinespace
\multicolumn{8}{@{}l}{\textit{Sex}} \\
National, all news & Women & Men & 495.1 & 491.4 & $+0.8\%$ & $-1.8$ to $+3.3$ & 1{,}766 and 1{,}246 \\
National, four topics & Women & Men & 505.8 & 486.7 & $+3.9\%$ & $-3.3$ to $+11.6$ & 237 and 162 \\
Museum adults, scenes & Women & Men & 331.3 & 322.8 & $+2.6\%$ & $+0.8$ to $+4.5$ & 1{,}936 and 2{,}716 \\
Six countries, scenes & Women & Men & 300.5 & 320.6 & $-6.3\%$ & $-9.1$ to $-3.4$ & 216 and 190 \\
University, videos & Women & Men & 386.6 & 391.9 & $-1.4\%$ & $-6.6$ to $+3.8$ & 151 and 171 \\
\addlinespace
\multicolumn{8}{@{}l}{\textit{Party}} \\
National, all news & Republicans & Democrats & 496.3 & 487.7 & $+1.8\%$ & $-0.9$ to $+4.4$ & 1{,}158 and 1{,}380 \\
National, four topics & Republicans & Democrats & 503.6 & 493.6 & $+2.0\%$ & $-4.8$ to $+8.7$ & 177 and 222 \\
\addlinespace
\multicolumn{8}{@{}l}{\textit{All characteristics together, viewers held out of the fit}} \\
National, all news & Predicted longest fifth & Predicted shortest fifth & 529.7 & 453.2 & $+16.9\%$ & $+12.3$ to $+21.6$ & 603 and 603 \\
\bottomrule
\end{tabular}}%
\resizebox{\textwidth}{!}{\usebox{\tabbox}}
\par\smallskip
{\scriptsize\justifying\settabnotew\noindent\parbox{\tabnotew}{\setlength{\emergencystretch}{2em}\textit{Notes.} The measure is how long a viewer's fixations last on average, first within a viewing and then over the viewer's viewings. In the public datasets it is the fixation duration the tracker reports, and in the national sample it is the interval defined in Methods. The difference is the first group's mean divided by the second group's mean, minus one, so a positive difference means the first group's fixations last longer, with a 95\%~interval from 2,000 bootstrap draws of viewers. National rows refer to the national sample of this study, with the average of each image and of each device class subtracted. They cover all 83~news images or one of the two image pools, the 48~images on four topics or the 35~images on immigration. The public datasets and the positions on the issues are described in Methods. A viewer leans toward a side when the viewer's mean position lies more than half a step from the middle of the scale, and the row below it takes viewers more than one step from the middle. Ideology rows give points of the 7-point scale. The last row sorts held-out viewers by the time per fixation that a regression on their characteristics predicts, and it compares the fifth predicted to look longest with the fifth predicted to look shortest.\par}}
\end{table}

\begin{table}[H]
\centering
\caption{Biases of a single prediction in the number of fixations and in the text reached.}
\label{tab:places_text}
\scriptsize
\setlength{\tabcolsep}{4pt}
\renewcommand{\arraystretch}{0.74}
\sbox{\tabbox}{\begin{tabular}{@{}lll rr r c r@{}}
\toprule
 & & & \multicolumn{2}{c}{Value} & & & \\
\cmidrule(lr){4-5}
Dataset and content & First group & Second group & First & Second & Difference & 95\% interval & Viewers \\
\midrule
\multicolumn{8}{@{}l}{\textit{Fixations in one viewing, by age}} \\
Museum visitors, scenes & 75 and older & 19 to 34 & 5.88 & 7.11 & $-17.3\%$ & $-$20.7 to $-$13.8 & 126 and 1{,}823 \\
Museum visitors, scenes & 3 to 6 & 19 to 34 & 5.88 & 7.11 & $-17.2\%$ & $-$19.2 to $-$15.2 & 378 and 1{,}823 \\
Museum visitors, scenes & 55 and older & 19 to 34 & 6.55 & 7.11 & $-7.8\%$ & $-$9.3 to $-$6.3 & 910 and 1{,}823 \\
Museum visitors, scenes & 9 and younger & 19 to 34 & 6.08 & 7.11 & $-14.5\%$ & $-$15.6 to $-$13.4 & 2{,}171 and 1{,}823 \\
Wilming and colleagues, scenes & Children & Young adults & 11.86 & 14.57 & $-18.6\%$ & $-$25.5 to $-$11.8 & 18 and 23 \\
Wilming and colleagues, scenes & Older adults & Young adults & 15.59 & 14.57 & $+7.0\%$ & $-$2.9 to $+$17.7 & 17 and 23 \\
Young viewers, videos & 5 to 7 & 14 to 17 & 2.82 & 2.92 & $-3.2\%$ & $-$6.5 to $+$0.8 & 558 and 252 \\
National, all news & 55 and older & 18 to 34 & 10.39 & 11.24 & $-7.5\%$ & $-$10.9 to $-$4.5 & 766 and 883 \\
National, four topics & 55 and older & 18 to 34 & 9.36 & 12.09 & $-22.6\%$ & $-$30.6 to $-$13.6 & 132 and 76 \\
\addlinespace
\multicolumn{8}{@{}l}{\textit{Fixations in one viewing, by race}} \\
National, all news & Black & White & 9.81 & 10.99 & $-10.8\%$ & $-$14.7 to $-$6.8 & 335 and 2{,}302 \\
University, videos & Black & White & 2.06 & 2.31 & $-10.6\%$ & $-$17.6 to $-$3.3 & 29 and 177 \\
University, videos & Hispanic & White & 2.24 & 2.31 & $-2.8\%$ & $-$7.6 to $+$1.8 & 80 and 177 \\
\addlinespace
\multicolumn{8}{@{}l}{\textit{Fixations in one viewing, by position on the issues}} \\
National, four topics & Leans Republican & Leans Democratic & 9.02 & 11.28 & $-20.0\%$ & $-$27.8 to $-$11.2 & 80 and 162 \\
National, four topics & Republican, over 1 step & Democratic, over 1 step & 8.90 & 11.30 & $-21.3\%$ & $-$31.6 to $-$9.3 & 45 and 132 \\
National, immigration & Leans Republican & Leans Democratic & 10.80 & 11.10 & $-2.7\%$ & $-$5.3 to $+$0.1 & 1{,}202 and 893 \\
\addlinespace
\multicolumn{8}{@{}l}{\textit{Fixations in one viewing, other axes}} \\
National, all news & Brackets 1 to 4 & Brackets 9 to 12 & 10.44 & 11.18 & $-6.6\%$ & $-$9.4 to $-$3.7 & 911 and 1{,}069 \\
National, all news & Levels 1 to 4 & Levels 5 to 7 & 10.69 & 10.96 & $-2.4\%$ & $-$5.0 to $+$0.1 & 1{,}561 and 1{,}450 \\
National, all news & Republicans & Democrats & 10.72 & 10.98 & $-2.3\%$ & $-$4.8 to $+$0.2 & 1{,}158 and 1{,}380 \\
National, all news & Most conservative & Most liberal & 10.51 & 11.49 & $-8.5\%$ & $-$13.4 to $-$3.6 & 306 and 384 \\
National, all news & Points 1 and 7 & Point 4 & 11.05 & 10.74 & $+2.9\%$ & $-$0.6 to $+$6.6 & 690 and 812 \\
National, all news & Women & Men & 10.76 & 10.90 & $-1.3\%$ & $-$3.7 to $+$1.2 & 1{,}766 and 1{,}246 \\
Museum adults, scenes & Women & Men & 7.00 & 7.15 & $-2.1\%$ & $-$3.1 to $-$1.0 & 1{,}936 and 2{,}716 \\
Six countries, scenes & Ghana and Turkey & Germany and Czechia & 9.58 & 10.21 & $-6.1\%$ & $-$11.2 to $-$0.5 & 141 and 146 \\
Six countries, scenes & Women & Men & 10.02 & 9.83 & $+2.0\%$ & $-$2.8 to $+$6.8 & 216 and 190 \\
Six countries, scenes & Without university & University & 10.14 & 9.81 & $+3.3\%$ & $-$1.6 to $+$8.5 & 158 and 250 \\
University, videos & Women & Men & 2.29 & 2.24 & $+2.0\%$ & $-$1.6 to $+$5.8 & 151 and 171 \\
\addlinespace
\multicolumn{8}{@{}l}{\textit{Share of gaze time on text, images that carry text}} \\
Museum visitors, scenes & 3 to 6 & 19 to 34 & 6.4 & 21.6 & $-70.3\%$ & $-$73.3 to $-$66.9 & 360 and 1{,}796 \\
Museum visitors, scenes & 75 and older & 19 to 34 & 16.3 & 21.6 & $-24.3\%$ & $-$32.8 to $-$16.2 & 121 and 1{,}796 \\
Museum visitors, scenes & 55 and older & 19 to 34 & 20.8 & 21.6 & $-3.5\%$ & $-$6.8 to $+$0.0 & 894 and 1{,}796 \\
Museum adults, scenes & Women & Men & 22.1 & 22.2 & $-0.6\%$ & $-$2.8 to $+$1.9 & 1{,}909 and 2{,}684 \\
National, all news & 55 and older & 18 to 34 & 1.01 & 1.05 & $-3.8\%$ & $-$21.1 to $+$18.2 & 763 and 877 \\
National, all news & Brackets 1 to 4 & Brackets 9 to 12 & 0.98 & 1.18 & $-17.2\%$ & $-$31.3 to $-$2.3 & 903 and 1{,}066 \\
National, all news & Black & White & 0.89 & 1.03 & $-13.9\%$ & $-$33.2 to $+$11.9 & 333 and 2{,}290 \\
National, all news & Leans Republican & Leans Democratic & 0.97 & 1.03 & $-5.4\%$ & $-$19.8 to $+$12.2 & 1{,}277 and 1{,}048 \\
National, all news & Women & Men & 1.05 & 1.04 & $+0.7\%$ & $-$13.7 to $+$16.9 & 1{,}758 and 1{,}240 \\
\addlinespace
\multicolumn{8}{@{}l}{\textit{Share of viewings in which the gaze reached the text}} \\
Museum visitors, scenes & 3 to 6 & 19 to 34 & 24.5 & 60.7 & $-59.6\%$ & $-$62.8 to $-$56.6 & 360 and 1{,}796 \\
Museum visitors, scenes & 75 and older & 19 to 34 & 43.4 & 60.7 & $-28.5\%$ & $-$35.6 to $-$21.8 & 121 and 1{,}796 \\
Museum visitors, scenes & 55 and older & 19 to 34 & 55.4 & 60.7 & $-8.8\%$ & $-$11.3 to $-$6.2 & 894 and 1{,}796 \\
Museum adults, scenes & Women & Men & 61.0 & 61.7 & $-1.2\%$ & $-$3.0 to $+$0.7 & 1{,}909 and 2{,}684 \\
National, all news & 55 and older & 18 to 34 & 5.7 & 6.1 & $-6.9\%$ & $-$19.9 to $+$8.9 & 763 and 877 \\
National, all news & Brackets 1 to 4 & Brackets 9 to 12 & 5.2 & 7.3 & $-29.4\%$ & $-$38.0 to $-$19.8 & 903 and 1{,}066 \\
National, all news & Black & White & 5.1 & 6.2 & $-17.0\%$ & $-$31.2 to $-$2.2 & 333 and 2{,}290 \\
National, all news & Leans Republican & Leans Democratic & 5.6 & 6.4 & $-12.3\%$ & $-$22.4 to $-$2.1 & 1{,}277 and 1{,}048 \\
National, all news & Women & Men & 6.0 & 6.3 & $-5.3\%$ & $-$14.9 to $+$5.9 & 1{,}758 and 1{,}240 \\
\addlinespace
\multicolumn{8}{@{}l}{\textit{Share of fixations inside the top quarter of the DeepGazeIIE map, percent}} \\
Museum visitors, scenes & 75 and older & 19 to 34 & 83.1 & 84.3 & $-1.5\%$ & $-$4.7 to $+$1.5 & 126 and 1{,}823 \\
Museum visitors, scenes & 3 to 6 & 19 to 34 & 84.4 & 84.3 & $+0.1\%$ & $-$1.9 to $+$2.0 & 378 and 1{,}823 \\
National, all news & 55 and older & 18 to 34 & 44.4 & 47.7 & $-6.8\%$ & $-$9.4 to $-$4.1 & 766 and 883 \\
National, all news & Black & White & 44.8 & 46.3 & $-3.4\%$ & $-$7.0 to $+$0.1 & 335 and 2{,}302 \\
National, four topics & Leans Republican & Leans Democratic & 46.7 & 48.9 & $-4.4\%$ & $-$12.4 to $+$4.2 & 80 and 162 \\
\addlinespace
\multicolumn{8}{@{}l}{\textit{All characteristics together, viewers held out of the fit, number of fixations}} \\
Museum visitors, scenes & Predicted highest fifth & Predicted lowest fifth & 7.37 & 6.04 & $+22.0\%$ & $+$20.7 to $+$23.6 & 2{,}373 and 2{,}373 \\
National, all news & Predicted highest fifth & Predicted lowest fifth & 11.70 & 10.09 & $+15.9\%$ & $+$11.6 to $+$20.6 & 603 and 603 \\
Six countries, scenes & Predicted highest fifth & Predicted lowest fifth & 9.88 & 9.97 & $-0.9\%$ & $-$8.4 to $+$6.6 & 82 and 82 \\
University, videos & Predicted highest fifth & Predicted lowest fifth & 2.34 & 2.18 & $+7.0\%$ & $+$0.8 to $+$14.1 & 65 and 65 \\
Young viewers, videos & Predicted highest fifth & Predicted lowest fifth & 2.90 & 2.82 & $+2.6\%$ & $-$1.0 to $+$6.2 & 384 and 384 \\
\bottomrule
\end{tabular}}%
\resizebox{\textwidth}{!}{\usebox{\tabbox}}
\par\vspace{1pt}
{\scriptsize\justifying\settabnotew\noindent\parbox{\tabnotew}{\setlength{\emergencystretch}{2em}\textit{Notes.} The number of fixations is counted in one viewing of one image, a fixation being a stop of the gaze of at least 80~ms within 3\%~of the frame, and averaged within each viewer, with the average of each image subtracted and, in the national sample, the average of each device class as well. In the two video datasets the measure is fixations per second, with the average of the video subtracted. The university rows come from GazeBase, and the young viewers from the Healthy Brain Network, counted on the left eye alone (Supplementary Information S1). The share of gaze time on text is the share of fixation time on lines that optical character recognition reads with a confidence of 0.8 or more, averaged over the images that carry text, 20 of the 40 museum scenes and 36 of the 83 news images. The reached rows give the share of those viewings in which at least one fixation fell on text. The four deep networks together place 13.6\% of their map on the museum text and 3.8\% on the news text. The difference is the first group's mean divided by the second group's, minus one, with a 95\% interval from 1,000 bootstrap draws of viewers, and from 2,000 draws in the last block. The last block sorts held-out viewers by the number of fixations that a ridge regression on their characteristics predicts, as in Extended Data Table~\ref{tab:rest_time_all}.\par}}
\end{table}

\begin{table}[H]
\centering
\caption{A network model of groups, scored on viewers held out of its training.}
\label{tab:group_network}
\scriptsize
\setlength{\tabcolsep}{3.5pt}
\renewcommand{\arraystretch}{0.92}
\sbox{\tabbox}{\begin{tabular}{@{}ll r cccc cc@{}}
\toprule
 & & & \multicolumn{4}{c}{Network on fixations} & \multicolumn{2}{c}{Network on the raw webcam stream} \\
\cmidrule(lr){4-7}\cmidrule(lr){8-9}
 & & & \multicolumn{3}{c}{One viewer} & & & \\
\cmidrule(lr){4-6}
First group & Second group & Viewers & All & Place only & Duration only & 20 viewers & One viewer & 20 viewers \\
\midrule
\multicolumn{9}{@{}l}{\textit{Museum visitors, scenes}} \\
3 to 9 & 18 to 34 & 436 and 423 & \textbf{0.835} (0.735) & \textbf{0.833} (0.718) & \textbf{0.746} (0.700) & \textbf{1.000} (0.972) & & \\
10 to 17 & 18 to 34 & 918 and 423 & \textbf{0.715} (0.675) & \textbf{0.715} (0.679) & \textbf{0.634} (0.602) & \textbf{0.966} (0.914) & & \\
35 to 54 & 18 to 34 & 418 and 423 & \textbf{0.601} (0.532) & \textbf{0.602} (0.559) & 0.531 (0.557) & \textbf{0.815} (0.702) & & \\
55 and older & 18 to 34 & 182 and 423 & \textbf{0.677} (0.649) & \textbf{0.633} (0.628) & \textbf{0.697} (0.688) & \textbf{0.904} (0.805) & & \\
women & men & 1{,}028 and 1{,}275 & \textbf{0.584} (0.576) & 0.583 (0.585) & 0.539 (0.540) & 0.791 (0.795) & & \\
other gender & men & 110 and 1{,}275 & \textbf{0.639} (0.636) & 0.638 (0.642) & 0.580 (0.595) & \textbf{0.948} (0.930) & & \\
other gender & women & 110 and 1{,}028 & \textbf{0.602} (0.584) & \textbf{0.595} (0.592) & \textbf{0.627} (0.585) & \textbf{0.879} (0.842) & & \\
English & German & 150 and 2{,}263 & 0.598 (0.609) & 0.597 (0.615) & 0.548 (0.566) & \textbf{0.906} (0.871) & & \\
\addlinespace
\multicolumn{9}{@{}l}{\textit{Students of six countries, scenes}} \\
women & men & 43 and 38 & 0.540 (0.635) & 0.546 (0.660) & 0.518 (0.651) &  & & \\
without university & university & 30 and 51 & 0.513 (0.734) & 0.412 (0.738) & 0.604 (0.753) &  & & \\
Ghana and Turkey & Germany and Czechia & 27 and 26 & \textbf{0.799} (0.782) & \textbf{0.781} (0.711) & 0.702 (0.781) &  & & \\
lower class and poverty & upper middle and upper class & 6 and 19 & 0.421 (0.851) & 0.474 (0.825) & 0.316 (0.781) &  & & \\
21 and younger & 23 and older & 37 and 27 & 0.604 (0.670) & 0.624 (0.627) & 0.506 (0.638) &  & & \\
\addlinespace
\multicolumn{9}{@{}l}{\textit{Viewers of Wilming and colleagues, scenes}} \\
children & young adults & 3 and 4 & 1.000 (1.000) & 1.000 (1.000) & 1.000 (1.000) &  & & \\
older adults & young adults & 3 and 4 & 0.667 (0.917) & 0.667 (0.750) & 0.917 (1.000) &  & & \\
children & older adults & 3 and 3 & 1.000 (1.000) & 0.889 (1.000) & 1.000 (1.000) &  & & \\
\addlinespace
\multicolumn{9}{@{}l}{\textit{National sample, news images}} \\
55 and older & 18 to 34 & 151 and 174 & 0.501 (0.591) & 0.504 (0.597) & 0.480 (0.605) & 0.269 (0.883) & \textbf{0.608} (0.592) & \textbf{0.882} (0.838) \\
35 to 54 & 18 to 34 & 277 and 174 & 0.515 (0.547) & 0.524 (0.548) & 0.486 (0.546) & 0.476 (0.756) & 0.517 (0.572) & 0.566 (0.827) \\
women & men & 353 and 249 & 0.497 (0.541) & 0.519 (0.538) & 0.497 (0.536) & 0.521 (0.640) & \textbf{0.535} (0.514) & 0.532 (0.606) \\
Black & White & 69 and 450 & 0.566 (0.638) & 0.540 (0.661) & 0.583 (0.618) & 0.812 (0.950) & \textbf{0.608} (0.603) & \textbf{0.888} (0.877) \\
Asian & White & 39 and 450 & 0.363 (0.660) & 0.369 (0.646) & 0.413 (0.649) &  & \textbf{0.621} (0.609) &  \\
Hispanic & not Hispanic & 55 and 547 & 0.574 (0.580) & 0.550 (0.583) & 0.569 (0.578) & 0.749 (0.800) & 0.512 (0.537) & 0.703 (0.720) \\
income brackets 1 to 4 & brackets 9 to 12 & 184 and 213 & 0.465 (0.593) & 0.453 (0.581) & 0.480 (0.569) & 0.449 (0.861) & 0.577 (0.581) & 0.727 (0.779) \\
education levels 1 to 4 & levels 5 to 7 & 321 and 281 & 0.507 (0.580) & 0.508 (0.571) & 0.503 (0.588) & 0.487 (0.741) & 0.520 (0.526) & 0.559 (0.665) \\
Republicans & Democrats & 235 and 269 & 0.531 (0.540) & 0.521 (0.543) & 0.522 (0.535) & 0.606 (0.709) & 0.518 (0.520) & \textbf{0.512} (0.508) \\
ideology 1 to 3, conservative & 5 to 7, liberal & 205 and 224 & \textbf{0.570} (0.549) & 0.536 (0.554) & \textbf{0.565} (0.545) & 0.548 (0.726) & 0.480 (0.560) & 0.415 (0.725) \\
ideology 1, 2, 6, 7, strong conviction & 3 to 5, moderate & 297 and 305 & 0.425 (0.609) & 0.451 (0.609) & 0.422 (0.609) & 0.183 (0.913) & 0.472 (0.526) & 0.402 (0.675) \\
thermometer gap 50 and more & 20 and less & 256 and 192 & 0.504 (0.569) & 0.515 (0.546) & 0.506 (0.584) & 0.481 (0.832) & 0.520 (0.529) & 0.495 (0.728) \\
Democratic thermometer, top quarter & bottom quarter & 141 and 145 & 0.527 (0.565) & 0.515 (0.569) & 0.558 (0.563) & 0.704 (0.746) & \textbf{0.590} (0.567) & 0.820 (0.836) \\
Republican thermometer, top quarter & bottom quarter & 155 and 141 & 0.529 (0.569) & 0.519 (0.576) & 0.500 (0.573) & \textbf{0.768} (0.743) & 0.496 (0.539) & 0.441 (0.600) \\
attachment to own party 81 and more & 60 and less & 33 and 25 & 0.545 (0.568) & 0.528 (0.590) & 0.560 (0.606) &  & 0.562 (0.613) &  \\
leans Republican, four topics & leans Democratic & 15 and 31 & 0.411 (0.628) & 0.361 (0.622) & 0.525 (0.654) &  & 0.551 (0.680) &  \\
leans Republican, immigration & leans Democratic & 237 and 167 & 0.475 (0.537) & 0.451 (0.537) & \textbf{0.536} (0.535) & 0.307 (0.716) & 0.514 (0.543) & 0.542 (0.686) \\
abortion item 1 or 2 & 4 or 5 & 130 and 395 & 0.484 (0.556) & 0.511 (0.577) & 0.449 (0.550) & 0.554 (0.744) & 0.485 (0.588) & 0.292 (0.796) \\
affirmative action item 1 or 2 & 4 or 5 & 195 and 256 & 0.544 (0.545) & \textbf{0.575} (0.532) & 0.468 (0.536) & 0.591 (0.652) & 0.496 (0.526) & 0.506 (0.549) \\
college access item 1 or 2 & 4 or 5 & 48 and 456 & 0.413 (0.584) & 0.418 (0.591) & 0.449 (0.562) & 0.276 (0.798) & 0.475 (0.576) & 0.258 (0.817) \\
gun permits item 1 or 2 & 4 or 5 & 57 and 27 & 0.527 (0.540) & 0.539 (0.554) & 0.495 (0.607) &  & 0.435 (0.594) &  \\
January 6 item 1 or 2 & 4 or 5 & 8 and 72 & 0.406 (0.764) & 0.436 (0.760) & 0.472 (0.809) &  & 0.675 (0.752) &  \\
immigration item 1 or 2 & 4 or 5 & 279 and 199 & 0.487 (0.559) & 0.468 (0.551) & 0.527 (0.552) & 0.512 (0.667) & 0.527 (0.531) & 0.619 (0.674) \\
same-sex marriage item 1 or 2 & 4 or 5 & 139 and 357 & 0.515 (0.576) & 0.513 (0.565) & 0.511 (0.562) & 0.404 (0.873) & 0.563 (0.589) & 0.702 (0.803) \\
\bottomrule
\end{tabular}}%
\resizebox{\textwidth}{!}{\usebox{\tabbox}}
\par\smallskip
{\scriptsize\justifying\noindent\parbox{\textwidth}{\setlength{\emergencystretch}{2em}\textit{Notes.} Each cell gives the area under the curve with which the network tells the first group from the second among viewers held out by person, for a single viewer and for audiences of 20 viewers drawn from each group. A viewer's score is how much more probable the network finds the viewer's recorded gaze when it is told the first group than when it is told the second, measured as a difference of log likelihoods, with every other trait set to unknown. The place-only and duration-only columns score a single viewer on where each next fixation lands and on how long it lasts. In parentheses is the highest value among 20 networks trained after the group labels of the training viewers were shuffled. Bold marks a value above all 20. The fixation network learns from the four corpora together, and the raw-stream network from the national sample alone (Methods, Supplementary Information S5). Audience cells are empty where a group has fewer than 40 held-out viewers. With 24 national contrasts, chance alone would let about one of them exceed all 20 shuffled networks. The raw-stream network exceeds them on two of the four contrasts named for it in development, an outcome chance would give about once in 80 runs, and on four of the other 20. Issue items run from 1, the conservative end, to 5; attachment to own party is a 0 to 100 rating from wave 4.\par}}
\end{table}

\newpage
\setcounter{table}{0}
\setcounter{figure}{0}
\renewcommand{\tablename}{Supplementary Table}
\renewcommand{\figurename}{Supplementary Fig.}

\section*{Supplementary Information}

\subsection*{S1. Detail of the group comparisons}

On the released maps the models read Black viewers less accurately than White viewers, by about a tenth of what the deep networks add over the center map, in six of seven specifications (Extended Data Table~\ref{tab:spec_grid}).

The age gap keeps its sign in every specification and every wave. Extremity compares people at either end of the ideology scale with those in the middle. Which party a viewer leans to leaves the score unchanged.

With a correction for four axes, the all-device race estimate of UNISAL loses significance.

The Taiwan contrast among the students reflects its different tracker (Extended Data Table~\ref{tab:public_models}). Fixation counts of the Healthy Brain Network use the left eye, because the release lists both eyes in one stream (Extended Data Table~\ref{tab:places_text}).

\subsection*{S2. The nine datasets}
\label{si:datasets}

Supplementary Table~\ref{tab:datasets} lists the nine datasets and the analyses each one feeds.

\begin{table}[H]
\centering
\caption{The nine datasets and what each one supplies.}
\label{tab:datasets}
\footnotesize
\setlength{\tabcolsep}{4pt}
\sbox{\tabbox}{\begin{tabular}{@{}>{\raggedright\arraybackslash}p{0.17\textwidth} >{\raggedright\arraybackslash}p{0.09\textwidth} >{\raggedright\arraybackslash}p{0.12\textwidth} >{\raggedright\arraybackslash}p{0.12\textwidth} >{\raggedright\arraybackslash}p{0.20\textwidth} >{\raggedright\arraybackslash}p{0.15\textwidth}@{}}
\toprule
Dataset & Viewers & Stimuli & Instrument & Attributes recorded & Analyses \\
\midrule
National sample (this study) & 2{,}993 & 83 news images for 5 s & webcam, WebGazer & age, gender, race, education, income, party, ideology, positions on four issues & all \\
Museum visitors, Linka et al. & 12{,}056 & 40 scenes for 3 s & Tobii 4C at 90 Hz & age, gender, language, glasses & accuracy, duration, places, text, readouts, groups, network, feed \\
Students of six countries, \v{C}en\v{e}k et al. & 408 & 70 scenes for 4 s & SMI RED 250 mobile & country, sex, education, class & accuracy, duration, places, readouts, groups, network \\
Wilming et al. & 58 & 249 scenes for 5 s & EyeLink 1000 at 500 Hz & age group & accuracy, duration, places, readouts, network \\
Museum, one image, Strauch et al. & 6{,}909 and 1{,}233 & 1 image for 10 s & Tobii 4C and Tobii Pro Fusion & age, gender & accuracy \\
News pages, S\"umer et al. & 25 & 119 news pages, read at the viewer's pace or for up to 10 s & Tobii Pro Spectrum at 600 Hz & age, gender & accuracy \\
Young viewers, Healthy Brain Network & 3{,}380 with fixations, 2{,}426 with age and sex & 4 short videos & SMI iView X RED-m at 120 Hz & age, sex & accuracy, duration, places, groups \\
University students, GazeBase & 322 & 2 videos of 60 s & EyeLink 1000 & ethnicity, gender, age & duration, places \\
MIT1003, Judd et al. & 15 & 1{,}003 scenes for 3 s, 250 used & laboratory tracker & unrecorded & accuracy, map predictors, measurement error \\
\bottomrule
\end{tabular}}%
\ifdim\wd\tabbox>\textwidth\resizebox{\textwidth}{!}{\usebox{\tabbox}}\else\usebox{\tabbox}\fi
\par\smallskip
{\scriptsize\justifying\settabnotew\noindent\parbox{\tabnotew}{\setlength{\emergencystretch}{2em}\textit{Notes.} Viewers are the people with fixations in each dataset's gaze file. In the one-image museum study, 4{,}117 and 1{,}174 of them report age and gender and enter the analyses. Six of the seven student samples used the SMI RED 250 mobile. The students' country enters the tables only as a contrast between eye trackers, since the Taiwan sample was recorded on a different instrument, and it stays out of every conclusion about social groups.\par}}
\end{table}

\subsection*{S3. The feed experiment}
Supplementary Table~\ref{tab:feed_budgets} gives every result of the feed experiment of Fig.~\ref{fig:feed}, and Supplementary Table~\ref{tab:feed_checks} the checks with other radii and with a fixed number of fixations.

\begin{table}[H]
\centering
\caption{The feed experiment at two feed cuts and over the whole viewing.}
\label{tab:feed_budgets}
\scriptsize
\resizebox{\textwidth}{!}{%
\begin{tabular}{@{}lll llllll@{}}
\toprule
Data & First group & Second group & Map's first place & Second place & Third place & Of the first five & Text & Fixations begun \\
\midrule
\multicolumn{9}{@{}l}{\textit{1.7 seconds, mobile feed}} \\
Webcam, news & 55 and older & 18 to 34 & $-17.0$ [$-22.0$, $-11.9$] & $-19.4$ [$-24.2$, $-14.6$] & $-9.4$ [$-14.2$, $-4.1$] & $-14.5$ [$-18.4$, $-10.5$] & $+0.4$ [$-16.9$, $+22.4$] & $-8.8$ [$-12.0$, $-5.3$] \\
Webcam, news & Black & White & $-13.5$ [$-20.2$, $-6.5$] & $-11.6$ [$-18.2$, $-4.2$] & $-13.4$ [$-19.2$, $-7.3$] & $-12.3$ [$-17.5$, $-6.9$] & $-43.5$ [$-63.2$, $-22.2$] & $-10.5$ [$-15.0$, $-6.1$] \\
Webcam, news & Income brackets 1 to 4 & brackets 9 to 12 & $-9.4$ [$-14.3$, $-4.2$] & $-10.3$ [$-14.9$, $-5.4$] & $-6.1$ [$-10.8$, $-1.2$] & $-7.5$ [$-11.2$, $-3.3$] & $-9.9$ [$-25.2$, $+9.5$] & $-6.7$ [$-9.9$, $-3.4$] \\
Webcam, four topics & Leans Republican & leans Democratic & $-14.8$ [$-30.8$, $+3.1$] & $-24.3$ [$-37.0$, $-9.1$] & $-8.5$ [$-23.2$, $+6.8$] & $-15.9$ [$-27.8$, $-2.4$] & $-21.8$ [$-46.5$, $+13.7$] & $-21.2$ [$-30.1$, $-12.0$] \\
Museum, scenes & 75 and older & 19 to 34 & $-13.5$ [$-18.9$, $-8.7$] & $-18.1$ [$-23.2$, $-13.3$] & $-19.1$ [$-24.8$, $-13.6$] & $-16.7$ [$-21.5$, $-11.9$] & $-25.2$ [$-32.7$, $-17.3$] & $-16.8$ [$-20.7$, $-12.7$] \\
Museum, scenes & 55 and older & 19 to 34 & $-4.9$ [$-6.7$, $-3.1$] & $-6.4$ [$-8.3$, $-4.3$] & $-9.4$ [$-11.6$, $-7.1$] & $-6.6$ [$-8.3$, $-4.8$] & $-7.6$ [$-10.5$, $-4.5$] & $-7.7$ [$-9.3$, $-6.1$] \\
Museum, scenes & 3 to 6 & 19 to 34 & $-12.4$ [$-15.1$, $-9.5$] & $-12.9$ [$-16.0$, $-9.8$] & $-21.0$ [$-24.0$, $-17.8$] & $-15.4$ [$-18.1$, $-12.8$] & $-60.0$ [$-63.4$, $-56.8$] & $-14.6$ [$-16.7$, $-12.2$] \\
\addlinespace
\multicolumn{9}{@{}l}{\textit{2.5 seconds, desktop feed}} \\
Webcam, news & 55 and older & 18 to 34 & $-15.7$ [$-20.2$, $-10.9$] & $-16.6$ [$-21.0$, $-12.0$] & $-8.9$ [$-13.5$, $-4.3$] & $-13.0$ [$-16.9$, $-9.0$] & $-11.1$ [$-24.9$, $+4.7$] & $-8.6$ [$-11.9$, $-5.3$] \\
Webcam, news & Black & White & $-11.5$ [$-17.5$, $-5.4$] & $-12.1$ [$-17.7$, $-6.0$] & $-13.2$ [$-18.3$, $-7.7$] & $-11.2$ [$-16.1$, $-6.4$] & $-38.0$ [$-54.7$, $-20.0$] & $-10.6$ [$-14.9$, $-6.1$] \\
Webcam, news & Income brackets 1 to 4 & brackets 9 to 12 & $-8.5$ [$-12.8$, $-3.5$] & $-9.0$ [$-13.4$, $-4.5$] & $-5.7$ [$-9.9$, $-1.3$] & $-7.0$ [$-10.6$, $-3.1$] & $-9.8$ [$-23.1$, $+6.0$] & $-6.7$ [$-9.7$, $-3.6$] \\
Webcam, four topics & Leans Republican & leans Democratic & $-10.0$ [$-26.0$, $+6.5$] & $-21.2$ [$-33.6$, $-8.7$] & $-8.0$ [$-20.3$, $+6.8$] & $-13.1$ [$-24.8$, $-1.4$] & $-12.8$ [$-34.2$, $+15.1$] & $-20.2$ [$-28.7$, $-11.6$] \\
Museum, scenes & 75 and older & 19 to 34 & $-13.0$ [$-17.4$, $-8.5$] & $-15.3$ [$-19.9$, $-10.7$] & $-20.5$ [$-25.8$, $-15.4$] & $-16.4$ [$-20.8$, $-12.3$] & $-25.1$ [$-31.7$, $-18.0$] & $-18.1$ [$-22.3$, $-14.1$] \\
Museum, scenes & 55 and older & 19 to 34 & $-4.1$ [$-5.8$, $-2.4$] & $-5.2$ [$-7.0$, $-3.5$] & $-8.3$ [$-10.4$, $-6.3$] & $-6.0$ [$-7.6$, $-4.4$] & $-8.2$ [$-10.8$, $-5.6$] & $-7.8$ [$-9.4$, $-6.1$] \\
Museum, scenes & 3 to 6 & 19 to 34 & $-11.4$ [$-14.1$, $-8.8$] & $-14.9$ [$-17.7$, $-11.9$] & $-24.0$ [$-27.0$, $-21.2$] & $-16.7$ [$-19.1$, $-14.5$] & $-56.9$ [$-59.8$, $-53.8$] & $-17.2$ [$-19.4$, $-14.9$] \\
\addlinespace
\multicolumn{9}{@{}l}{\textit{Whole viewing, 5 s on the webcam and 3 s in the museum}} \\
Webcam, news & 55 and older & 18 to 34 & $-13.9$ [$-17.7$, $-10.1$] & $-13.4$ [$-17.3$, $-9.7$] & $-8.1$ [$-11.8$, $-4.3$] & $-11.4$ [$-14.4$, $-8.0$] & $-10.0$ [$-20.6$, $+1.5$] & $-8.1$ [$-11.4$, $-4.7$] \\
Webcam, news & Black & White & $-12.3$ [$-17.1$, $-7.2$] & $-9.0$ [$-14.2$, $-3.6$] & $-13.5$ [$-18.3$, $-8.6$] & $-11.0$ [$-15.2$, $-6.9$] & $-26.5$ [$-39.4$, $-12.6$] & $-11.3$ [$-15.5$, $-7.3$] \\
Webcam, news & Income brackets 1 to 4 & brackets 9 to 12 & $-6.5$ [$-10.2$, $-2.7$] & $-7.7$ [$-11.4$, $-3.9$] & $-6.1$ [$-9.6$, $-2.6$] & $-6.0$ [$-9.1$, $-2.7$] & $-13.9$ [$-23.0$, $-3.6$] & $-7.0$ [$-10.1$, $-3.8$] \\
Webcam, four topics & Leans Republican & leans Democratic & $-9.2$ [$-21.9$, $+4.9$] & $-10.6$ [$-21.0$, $+0.5$] & $-8.4$ [$-19.3$, $+4.6$] & $-9.9$ [$-20.3$, $+1.9$] & $+2.1$ [$-17.1$, $+23.6$] & $-20.1$ [$-28.4$, $-11.2$] \\
Museum, scenes & 75 and older & 19 to 34 & $-11.8$ [$-16.4$, $-7.3$] & $-14.5$ [$-19.0$, $-10.5$] & $-20.0$ [$-25.4$, $-15.0$] & $-15.4$ [$-19.8$, $-11.5$] & $-26.0$ [$-32.6$, $-19.7$] & $-18.8$ [$-22.9$, $-14.5$] \\
Museum, scenes & 55 and older & 19 to 34 & $-3.6$ [$-5.1$, $-2.0$] & $-4.8$ [$-6.6$, $-3.0$] & $-7.9$ [$-10.0$, $-5.9$] & $-5.6$ [$-7.4$, $-4.0$] & $-8.6$ [$-10.9$, $-6.2$] & $-8.0$ [$-9.6$, $-6.3$] \\
Museum, scenes & 3 to 6 & 19 to 34 & $-10.9$ [$-13.4$, $-8.6$] & $-15.2$ [$-17.9$, $-12.6$] & $-24.1$ [$-27.1$, $-21.2$] & $-16.9$ [$-19.2$, $-14.6$] & $-54.8$ [$-57.8$, $-51.6$] & $-18.5$ [$-20.9$, $-16.2$] \\
\addlinespace
\bottomrule
\end{tabular}}
\par\smallskip
{\scriptsize\justifying \textit{Notes.} Each cell is the first group's value minus the second group's, in percent of the second group's, with a 95\% interval from 2,000 bootstrap draws of people. The places are those the UNISAL map ranks first, second and third on the image, and the count covers the first five. A place is reached when a fixation that begins before the cut lands within 0.1 of the image width of it. Text is reached when such a fixation falls on a line of text. Fixations begun counts the fixations that start before the cut. What is typical for each image and, on the webcam, for each type of device is removed first (Methods).\par}
\end{table}

\begin{table}[H]
\centering
\caption{The place rule at three radii, and the first fixations whatever time they take.}
\label{tab:feed_checks}
\scriptsize
\resizebox{\textwidth}{!}{%
\begin{tabular}{@{}lll r lllll@{}}
\toprule
 & & & Time per & \multicolumn{3}{c}{Of the first five places within 1.7 s, radius} & \multicolumn{2}{c}{Of the first five places, any time} \\
\cmidrule(lr){5-7}\cmidrule(lr){8-9}
Data & First group & Second group & fixation & 0.075 & 0.10 & 0.125 & First three fixations & First five fixations \\
\midrule
Webcam, news & 55 and older & 18 to 34 & $+5.8$\% & $-17.3$ [$-21.9$, $-12.6$] & $-14.5$ [$-18.4$, $-10.5$] & $-13.2$ [$-16.9$, $-9.6$] & $-10.0$ [$-13.2$, $-6.7$] & $-9.8$ [$-12.8$, $-6.7$] \\
Webcam, news & Black & White & $+12.0$\% & $-14.1$ [$-20.0$, $-7.9$] & $-12.3$ [$-17.5$, $-6.9$] & $-10.9$ [$-15.5$, $-6.1$] & $-4.4$ [$-8.8$, $-0.1$] & $-5.8$ [$-10.0$, $-1.9$] \\
Webcam, news & Income brackets 1 to 4 & brackets 9 to 12 & $+7.7$\% & $-10.7$ [$-15.1$, $-5.9$] & $-7.5$ [$-11.2$, $-3.3$] & $-6.9$ [$-10.3$, $-3.2$] & $-3.7$ [$-7.0$, $-0.5$] & $-4.2$ [$-7.2$, $-1.0$] \\
Webcam, four topics & Leans Republican & leans Democratic & $+19.8$\% & $-17.2$ [$-32.3$, $-2.0$] & $-15.9$ [$-27.8$, $-2.4$] & $-14.9$ [$-25.9$, $-2.2$] & $-2.2$ [$-12.8$, $+9.6$] & $-4.8$ [$-15.3$, $+6.0$] \\
Museum, scenes & 75 and older & 19 to 34 & $+22.5$\% & $-20.0$ [$-25.4$, $-15.0$] & $-16.7$ [$-21.5$, $-11.9$] & $-13.5$ [$-18.0$, $-9.5$] & $-8.4$ [$-12.5$, $-4.3$] & $-11.3$ [$-15.3$, $-7.3$] \\
Museum, scenes & 55 and older & 19 to 34 &  & $-8.0$ [$-10.0$, $-6.0$] & $-6.6$ [$-8.3$, $-4.8$] & $-4.8$ [$-6.4$, $-3.2$] & $-2.3$ [$-3.7$, $-0.7$] & $-3.6$ [$-5.1$, $-2.0$] \\
Museum, scenes & 3 to 6 & 19 to 34 & $+18.5$\% & $-17.7$ [$-20.8$, $-14.9$] & $-15.4$ [$-18.1$, $-12.8$] & $-13.3$ [$-15.6$, $-11.0$] & $-8.3$ [$-10.7$, $-6.1$] & $-12.1$ [$-14.5$, $-9.8$] \\
\bottomrule
\end{tabular}}
\par\smallskip
{\scriptsize\justifying \textit{Notes.} Time per fixation is the difference of Table~\ref{tab:rest_time}. That table has no row for museum visitors aged 55 and older. Other cells as in Supplementary Table~\ref{tab:feed_budgets}.\par}
\end{table}

\subsection*{S4. Recording conditions and the model of groups}
The model of groups might read a person's group from the recording. A classifier that sees only the device identifies audiences of 20 people by income with 0.88 and by education with 0.89. One that sees only the wave of recruitment reaches 0.95 for income, 0.93 for age and 0.87 for education, because the waves recruited different people. The duration model refitted within image and device reaches 0.99 for age and 0.94 for income, about what the wave alone gives for income (Supplementary Table~\ref{tab:conditions}). With both audiences drawn from the same wave, as in Extended Data Table~\ref{tab:group_model}b, audiences of 10 are told apart by age with 0.86 and by income with 0.63.

\begin{table}[H]
\centering
\caption{Recording conditions and the model of groups, national sample.}
\label{tab:conditions}
\scriptsize
\resizebox{\textwidth}{!}{%
\begin{tabular}{@{}ll rl rl rl rl rl rl@{}}
\toprule
 & & \multicolumn{2}{c}{Device class only} & \multicolumn{2}{c}{Recruitment wave only} & \multicolumn{2}{c}{Duration, deviations} & \multicolumn{2}{c}{Places, deviations} & \multicolumn{2}{c}{Duration, full model} & \multicolumn{2}{c}{Places, full model} \\
\cmidrule(lr){3-4}\cmidrule(lr){5-6}\cmidrule(lr){7-8}\cmidrule(lr){9-10}\cmidrule(lr){11-12}\cmidrule(lr){13-14}
First group & Second group & One & 20 viewers & One & 20 viewers & One & 20 viewers & One & 20 viewers & One & 20 viewers & One & 20 viewers \\
\midrule
18 to 34 & 55 and older & 0.54 & 0.73 [0.63, 0.82] & 0.62 & 0.93 [0.88, 0.96] & 0.68 & 0.99 [0.97, 0.99] & 0.56 & 0.76 [0.67, 0.84] & 0.68 & 0.99 [0.97, 0.99] & 0.60 & 0.89 \\
Black & White & 0.48 & 0.39 [0.25, 0.52] & 0.54 & 0.73 [0.60, 0.85] & 0.58 & 0.82 [0.71, 0.90] & 0.50 & 0.53 [0.39, 0.69] & 0.58 & 0.82 [0.72, 0.89] & 0.53 & 0.72 \\
income brackets 1 to 4 & brackets 9 to 12 & 0.57 & 0.88 [0.83, 0.94] & 0.62 & 0.95 [0.92, 0.98] & 0.63 & 0.94 [0.89, 0.96] & 0.55 & 0.73 [0.64, 0.82] & 0.63 & 0.94 [0.89, 0.96] & 0.59 & 0.87 \\
education levels 1 to 4 & levels 5 to 7 & 0.57 & 0.89 [0.83, 0.92] & 0.59 & 0.87 [0.82, 0.92] & 0.58 & 0.85 [0.79, 0.89] & 0.52 & 0.61 [0.52, 0.70] & 0.59 & 0.87 [0.82, 0.92] & 0.53 & 0.67 \\
Democrats & Republicans & 0.50 & 0.51 [0.40, 0.61] & 0.49 & 0.42 [0.33, 0.53] & 0.49 & 0.42 [0.34, 0.52] & 0.49 & 0.42 [0.34, 0.51] & 0.49 & 0.44 [0.36, 0.56] & 0.49 & 0.46 \\
ideology 1 to 3 & 5 to 7 & 0.53 & 0.64 [0.55, 0.74] & 0.51 & 0.60 [0.50, 0.70] & 0.49 & 0.44 [0.35, 0.54] & 0.49 & 0.48 [0.36, 0.57] & 0.49 & 0.46 [0.37, 0.58] & 0.50 & 0.48 \\
strong conviction & moderate & 0.53 & 0.65 [0.56, 0.73] & 0.55 & 0.76 [0.68, 0.83] & 0.55 & 0.75 [0.68, 0.81] & 0.50 & 0.54 [0.44, 0.61] & 0.55 & 0.73 [0.64, 0.80] & 0.51 & 0.54 \\
leans Republican, four topics & leans Democratic & 0.50 & 0.50 [0.50, 0.51] & 0.50 & 0.50 [0.50, 0.50] & 0.60 & 0.86 [0.53, 0.97] & 0.57 & 0.77 [0.42, 0.92] & 0.59 & 0.84 [0.51, 0.96] & 0.59 & 0.87 \\
leans Republican, immigration & leans Democratic & 0.50 & 0.49 [0.41, 0.62] & 0.50 & 0.53 [0.42, 0.61] & 0.51 & 0.54 [0.44, 0.64] & 0.51 & 0.55 [0.45, 0.65] & 0.50 & 0.53 [0.39, 0.61] & 0.52 & 0.63 \\
thermometer gap 50 and more & 20 and less & 0.50 & 0.52 [0.42, 0.64] & 0.59 & 0.86 [0.78, 0.91] & 0.57 & 0.82 [0.73, 0.88] & 0.53 & 0.65 [0.54, 0.74] & 0.58 & 0.83 [0.74, 0.89] & 0.54 & 0.71 \\
\bottomrule
\end{tabular}}
\par\smallskip
{\scriptsize\justifying \textit{Notes.} Area under the curve for one held-out person, averaged over the three repetitions of five splits by person, and for audiences of 20 people drawn from each group across all waves, with a 95\% interval from 200 bootstrap resamples of the people. Device class only sees the share of a person's viewings on each type of device. Recruitment wave only sees the wave in which the person was recruited. The deviation models see the quantities of Methods only as differences from the training viewers of the same image on the same type of device, together with the agreement of the person's time and places with the map of each group. The full models are those of Extended Data Table~\ref{tab:group_model}.\par}
\end{table}

\subsection*{S5. The network model of groups}
\label{si:network}
The network describes each image with DINOv2, a self-supervised image encoder \autocite{oquab2024dinov2}, together with the DeepGazeIIE and UNISAL maps. It describes each person with numbers it learns for each known trait and for the recording conditions, and these numbers change how it reads the image. From this it draws a map for the next fixation and a duration. During training each trait is hidden at random, so the network also predicts for people whose group is unknown. A second network reads the raw webcam stream directly, and there the group can change only the timing of gaze.

\end{document}